\documentclass[runningheads]{llncs}
\usepackage{graphicx}

\usepackage{tikz}
\usepackage{comment}
\usepackage{amsmath,amssymb} 
\usepackage{color}

\usepackage{enumitem}
\usepackage{tablefootnote}
\usepackage{multirow}
\usepackage{capt-of}
\usepackage{fixltx2e}
\usepackage{floatrow}
\usepackage{microtype}
\usepackage{pifont}

\usepackage[colorlinks]{hyperref}

\def\etal{\emph{et al.}}
\def\etc{\emph{etc}}

\begin{document}
	\pagestyle{headings}
	\mainmatter
	\def\ECCVSubNumber{2270}  
	
	\title{Prior-based  Domain Adaptive Object Detection for Hazy and Rainy Conditions} 

	\titlerunning{Prior-based Domain Adaptive Object Detection under Haze \& Rain}
	%
	\author{Vishwanath A. Sindagi\thanks{equal contribution} \and
		Poojan Oza\textsuperscript{$\star$} \and
		Rajeev Yasarla\and
		Vishal M. Patel}
	\authorrunning{Sindagi  et al.}
	%
	\institute{	Department of Electrical and Computer Engineering,\\
		Johns Hopkins University, 3400 N. Charles St, Baltimore, MD 21218, USA
		\email{	{\tt\small \{vishwanathsindagi,poza2,ryasarl1,vpatel36\}@jhu.edu}}\\
	}
	\maketitle
	
	\begin{abstract}

		Adverse weather conditions such as haze and rain corrupt the quality of captured images, which cause detection networks trained on clean images to perform poorly on these images. To address this issue, we propose an unsupervised prior-based domain adversarial object detection framework for adapting the detectors to hazy and rainy conditions.  In particular, we use weather-specific prior knowledge obtained using the principles of image formation to define a novel prior-adversarial loss. The prior-adversarial loss used to train the adaptation process aims to reduce the weather-specific information in the features, thereby mitigating the effects of weather on the detection performance. Additionally, we introduce a set of residual feature recovery blocks in the object detection pipeline to de-distort the feature space, resulting in further improvements. Evaluations performed on various datasets (Foggy-Cityscapes, Rainy-Cityscapes, RTTS and UFDD) for rainy and hazy conditions demonstrates the effectiveness of the proposed approach.
		
	\end{abstract}

	\section{Introduction}\label{sec:introduction}

	Object detection \cite{viola2001rapid,felzenszwalb2010object,girshick2014rich,girshick2015fast,liu2016ssd,ren2015faster,sindagi2019dafe} is an extensively researched topic in the literature. Despite the success of deep learning based detectors on benchmark datasets  \cite{everingham2010pascal,deng2009imagenet,geiger2013vision,lin2014microsoft}, they  have limited abilities in generalizing to several  practical conditions such as adverse weather due to the domain shift in the input images.
	\begin{figure}[ht!]
		\begin{center}
			\includegraphics[width=.49\linewidth]{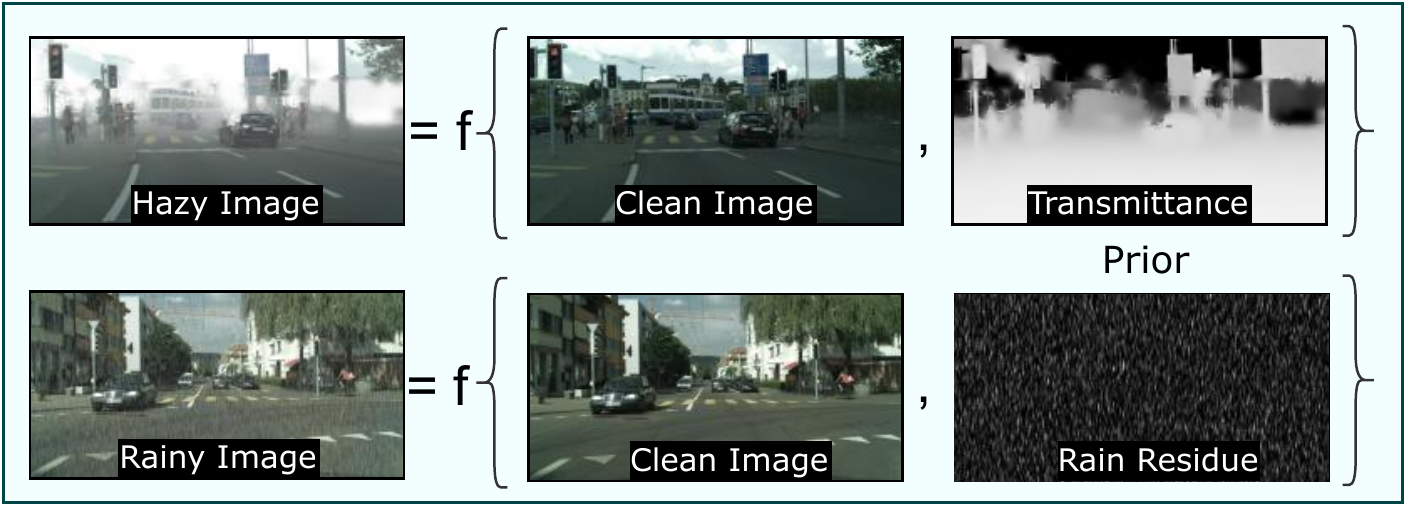}
			\includegraphics[width=.49\linewidth]{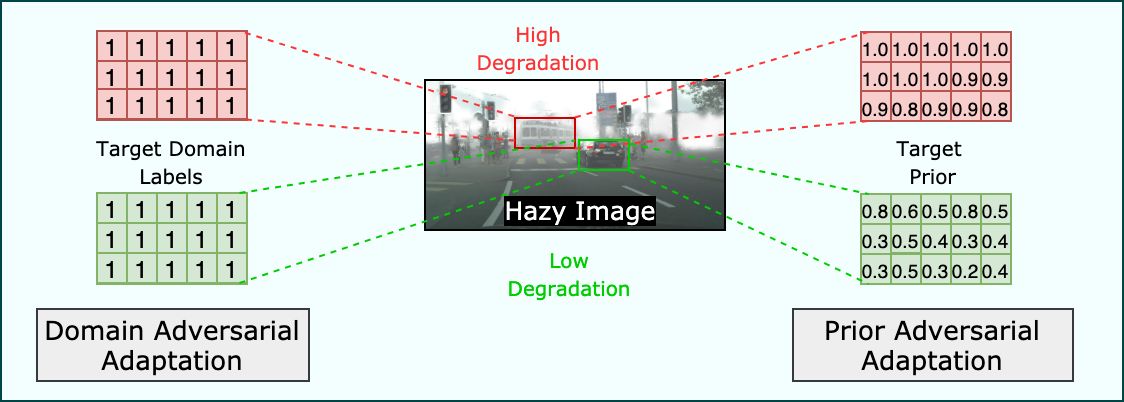}
			(a)\hskip150pt(b)
		\end{center}
		\vskip -18pt \caption{(a) Weather conditions such as rain and haze can be mathematically modeled as function of clean image and the weather-specific prior. We use this weather-specific prior to define a novel prior-adversarial loss for adapting detectors to adverse weather. (b) Existing domain adaptation approaches use constant target domain label for the entire image irrespective of the amount of degradation. Our method uses spatially-varying priors that are directly correlated to the amount of degradations.}
		\label{fig:img1}
	\end{figure}
	One approach to solve this issue is to undo the effects of weather conditions by pre-processing the images using existing methods like image dehazing \cite{fattal2008single,he2011single,zhang2018densely} and/or deraining \cite{Authors16,Authors17e,Authors18}. However,  these approaches usually involve complicated networks and need to be trained separately with pixel-level supervision. Moreover, as noted in \cite{Sakaridis2018SemanticFS}, these methods additionally involve certain post-processing like gamma correction, which still results in a domain shift, thus prohibiting such approaches from achieving the optimal performance. Like \cite{Sakaridis2018SemanticFS}, we   observed minimal   improvements in the detection performance when we used  state-of-the-art dehaze/derain methods  as pre-processing step before detection (see Sec. \ref{sec:experiments} ). Furthermore, this additional pre-processing would result in increased computational overhead at inference, which is not be preferable in resource-constrained/real-time applications. Another approach would be to re-train the detectors on datasets that include these adverse conditions. However, creating these datasets often comes with high annotation/labeling cost \cite{sindagi2019pushing}.
	
	Recently, a few methods \cite{Chen2018DomainAF,Shan2018,Saito2018StrongWeakDA} have attempted to overcome this problem by viewing object detection in adverse weather conditions as an unsupervised domain adaptation task.  These approaches consider that the images captured under adverse conditions (target images) suffer from a distribution shift \cite{Chen2018DomainAF,gopalan2011domain} as compared to the images on which the detectors are trained (source images). It is assumed that the source images are fully annotated while the target images (with weather-based degradations) are not annotated. They propose different techniques to align the target features with the source features, while training on the source images. These methods are inherently limited in their approach since they employ only the principles of domain adaptation and neglect additional information that is readily available in the case of weather-based degradations.  
	
	We consider the following observations about weather-based degradations which have been ignored in the earlier work. \textbf{\textit{(i)}} Images captured under weather conditions (such as haze and rain) can be mathematically modeled  (see Fig. \ref{fig:img1}(a), Eq. \ref{eq:haze} and \ref{eq:rain}). For example, a hazy image is modeled by a superposition of a clean image (attenuated by transmission map) and atmospheric light \cite{fattal2008single,he2011single}.  Similarly, a rainy image  is   modeled as a superposition of a clean image and rain residue \cite{Authors16,Authors18,Authors17e} (see Fig. \ref{fig:img1}(a)). In other words, a weather-affected image contains weather specific information (which we refer to as prior) - transmission map in the case of hazy images and rain residue in the case of rainy images. These weather-specific information/prior cause degradations in the feature space resulting in poor detection performance. Hence, in order to reduce the degradations in the features, it is crucial to make the features weather-invariant by eliminating the weather-specific priors from the features. \textbf{\textit{(ii)}} Further,  it is important to note that the weather-based degradations are spatially varying and, hence do not affect the features equally at all spatial locations. Since, existing domain-adaptive detection approaches \cite{Chen2018DomainAF,Shan2018,Saito2018StrongWeakDA}  label all the locations entirely either as target, they  assume that the entire image has undergone constant degradation and all spatial locations are equally affected (see Fig. \ref{fig:img1}(b)). This can potentially lead to incorrect alignment, especially in the regions of images where the degradations are minimal.
	
	Motivated by these observations, we define a novel prior-adversarial loss that uses additional knowledge about the target domain (weather-affected images) for aligning the source and target features. Specifically, the proposed loss is used to train a prior estimation network to predict weather-specific prior from the features in the main branch, while simultaneously minimizing the weather-specific information present in the features. This  results in weather-invariant features in the main branch, hence, mitigating the effects of weather. Additionally, the proposed use of prior information in the loss function results in spatially varying loss that is directly correlated to the amount of degradation (as shown in Fig. \ref{fig:img1}(b)). Hence, the use of prior can help avoid incorrect alignment.  
	
	Finally, considering that the weather-based degradations cause distortions in the feature space, we introduce a set of residual feature recovery blocks in the object detection pipeline to de-distort the features.  These blocks,  inspired by residual transfer framework proposed in \cite{he2016deep}, result in further improvements. 
	
	We perform  extensive evaluations on different datasets such as  Foggy-Cityscapes \cite{Sakaridis2018SemanticFS}, RTTS \cite{Li2018BenchmarkingSD} and UFDD \cite{nada2018pushing}. Additionally, we create a Rainy-Cityscapes dataset for evaluating the performance different detection methods on rainy conditions. Various experiments demonstrate that the proposed method is able to outperform the existing methods on all the datasets.

	\section{Related Work}
	\label{sec:related_work} 
	
	\noindent{\bf{Object detection}}: Object detection is one of the most researched topics in computer vision. Typical solutions for this problem have evolved from approaches involving sliding window based classification \cite{viola2001rapid,dalal2005histograms} to the latest anchor-based convolutional neural network approaches \cite{ren2015faster,redmon2016you,liu2016ssd}.  Ren \etal \cite{ren2015faster} pioneered the popular two stage Faster-RCNN approach. Several works have proposed single stage frameworks such as SSD \cite{liu2016ssd}, YOLO \cite{redmon2016you} \etc, that directly predict the object labels and bounding box co-ordinates. Following the previous work \cite{Chen2018DomainAF,Shan2018,Saito2018StrongWeakDA,kim2019diversify,khodabandeh2019robust} we use Faster-RCNN as our base model.\\
	\noindent \textbf{Unsupervised Domain Adaptation}: Unsupervised domain adaptation is defined as aligning domains having distinct distributions, namely source and target \cite{patel2015visual}. It is assumed that images in the source dataset are available with annotations, while no annotation information is provided for the target images. Some of the recently proposed methods for unsupervised domain adaptation include feature distribution alignment \cite{tzeng2017adversarial,ganin2014unsupervised,shu2018dirt,saito2018maximum,sindagi2017domain,sindagi2020learning,Yasarla_2020_CVPR}, residual transfer \cite{long2016unsupervised,long2017deep}, and image-to-image translation approaches \cite{hu2018duplex,murez2018image,hoffman2017cycada,sankaranarayanan2018generate,abavisani2018adversarial,abavisani2016domain,perera2018in2i}. In feature distribution alignment, an adversarial objective is utilized to learn domain-invariant features. Typically, these methods are implemented using a gradient reversal layer, where feature generator and domain classifier play an adversarial game to generate the target features that are aligned with the source feature distribution. Most of the research in unsupervised domain adaptation has focused on classification/segmentation problems and other tasks such as object detection are relatively unexplored.\\
	\noindent{\bf{Domain-adaptive object detection in adverse conditions}}: Compared to the problem of general detection, detection in adverse weather conditions is relatively less explored. Existing methods, \cite{Chen2018DomainAF,Shan2018,Saito2018StrongWeakDA,kim2019diversify} have attempted to address this task from a domain adaptation perspective. Chen \etal \cite{Chen2018DomainAF} assumed that the adversarial weather conditions result in domain shift, and they overcome this by proposing a domain adaptive Faster-RCNN approach that tackles domain shift on image-level and instance-level. Following the similar argument of domain shift,  Shan \etal \cite{Shan2018} proposed to perform joint adaptation at image level using the Cycle-GAN framework \cite{zhu2017unpaired} and at feature level using conventional domain adaptation losses. Saito \etal \cite{Saito2018StrongWeakDA}	argued that strong alignment of the features at global level might hurt the detection performance. Hence, they proposed a method which employs strong alignment of the local features and weak alignment of the global features.   Kim \etal \cite{kim2019diversify} diversified the labeled data, followed by adversarial learning with the help of multi-domain discriminators. Cai \etal \cite{cai2019exploring} addressed this problem in the semi-supervised setting using mean teacher framework. Zhu \etal \cite{zhu2019adapting} proposed region mining and region-level alignment in order to correctly align the source and target features. Roychowdhury \etal \cite{roychowdhury2019automatic} adapted detectors to a new domain assuming availability of large number of video data from the target domain. These video data are used to generate pseudo-labels for the target set, which are further employed to train the network.  Most recently, Khodabandeh \etal \cite{khodabandeh2019robust} formulated the domain adaptation training with noisy labels. Specifically, the model is trained on the target domain using a set of noisy bounding boxes that are obtained by a detection model trained only in the source domain.

	\section{Proposed Method}\label{sec:proposed_method}
	
\begin{figure*}[t!]
		\begin{center}
			\includegraphics[width=0.8\linewidth]{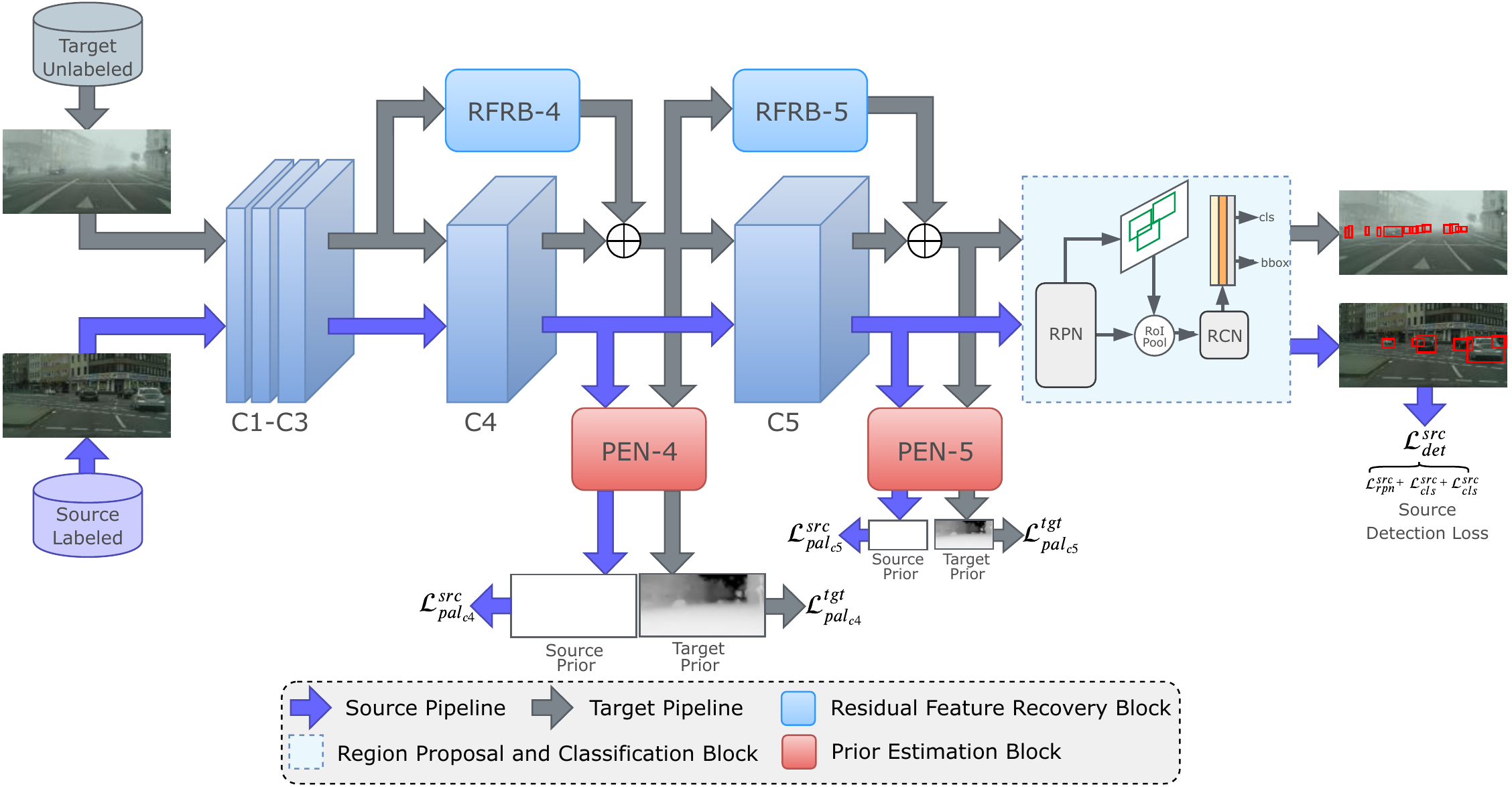}
		\end{center}
		\vskip -15.0pt \caption{Overview of the proposed adaptation method. We apply proposed prior adversarial loss at multiple scale of the network. The prior adversarial loss is supervised by source and target prior of respective sizes. For source pipeline, additional supervision is provided by detection loss. For target pipeline, feed-forward through the detection network is modified by the residual feature recovery blocks.}
		\label{fig:arch}
	\end{figure*}

	
	We assume that labeled clean data ($\{x_i^s,y_i^s\}_{i=1}^{n_s}$) from the source domain ($\mathcal{S}$)  
	and unlabeled weather-affected data from the target domain ($\mathcal{T}$)  are available. Here,  ${y}_i^s$ refers to all bounding box annotations and respective category label for the corresponding clean image ${x}_i^s$,  ${x}_i^t$ refers to the weather-affected image, $n_s$ is the total number of samples in the source domain  ($\mathcal{S}$)   and  $n_t$ is the total number of samples in the target domain  ($\mathcal{T}$). Our goal is to utilize the available information in both source and target domains to learn a network that lessens the effect of weather-based conditions on the detector.  The proposed method contains three network modules -- detection network, prior estimation network (PEN) and residual feature recovery block (RFRB).  Fig. \ref{fig:arch} gives an overview of the proposed model.  During source training, a source image (clean image) is passed to the detection network and the weights are learned by minimizing the detection loss, as shown in Fig. \ref{fig:arch} with the source pipeline.
	For target training, a target image (weather-affected image) is forwarded through the network as shown in Fig. \ref{fig:arch} by the target pipeline.
	As discussed earlier, weather-based degradations cause distortions in the feature space for the target images.
	In an attempt to de-distort these features, we introduce a set of residual feature recovery blocks in the target pipeline as shown in Fig. \ref{fig:arch}.
	This model is inspired from residual transfer framework proposed in \cite{long2016unsupervised} and is used to model residual features.
	The proposed PEN aids the detection network in adapting to the target domain by providing feedback through adversarial training using the proposed prior adversarial loss.
	In the following subsections, we briefly review the backbone network, followed by a  discussion on the proposed prior-adversarial loss and residual feature recovery blocks.\\
	
	\subsection{Detection Network}\label{subsec:detection_network}
	
	Following the existing domain adaptive detection approaches \cite{Chen2018DomainAF,Shan2018,Saito2018StrongWeakDA}, we base our method on the Faster-RCNN \cite{ren2015faster} framework.
	Faster-RCNN is among the first end-to-end CNN-based object detection methods and uses anchor-based strategy to perform detection and classification.
	For this paper we decompose the Faster-RCNN network into three network modules: feature extractor network ($\mathcal{F}$), region proposal network (RPN) stage and region classification network (RCN).
	The arrangement of these modules are shown in the Fig. \ref{fig:arch} with VGG model architecture as base network.
	Here, the feature extractor network consists of first five conv blocks of VGG and region classification network module is composed of fully connected layers of VGG.
	The region proposal network uses output of feature extractor network to generate a set of candidate object regions in a class agnostic way.
	Features corresponding to these candidates are pooled from the feature extractor and are forwarded through the region classification network to get the object classifications and bounding box refinements.
	Since we have access to the source domain images and their corresponding ground truth, these networks are trained to perform detection on the source domain by minimizing the following loss function,
	
	\vskip -19.0 pt
	\begin{align}\label{eq:detection_loss}
	&\underset{\mathcal{F}, \ \mathcal{G}}{\text{min}} \ \ \ \ \mathcal{L}^{src}_{det} \ ,\;\;\;\text{where} \\  
	\mathcal{L}^{src}_{det} \ &= \ \mathcal{L}^{src}_{rpn} + \mathcal{L}^{src}_{bbox} + \mathcal{L}^{src}_{rcn}.
	\end{align}
	Here, $\mathcal{G}$ represents both region proposal and region classification networks, $\mathcal{L}^{src}_{rpn}$ denotes the region proposal loss, $\mathcal{L}^{src}_{bbox}$ denotes the bounding-box regression loss and $\mathcal{L}^{src}_{rcn}$ denotes the region classification loss.  The details of these individual loss components can be found in \cite{ren2015faster}.

	\subsection{Prior-adversarial Training}\label{subsec:pal}
	As discussed earlier, weather-affected images, contain domain specific information.
	These images typically follow mathematical models of image degradation (see Fig. \ref{fig:img1}(a), Eq. \ref{eq:haze} and Eq. \ref{eq:rain}).
	We refer to this domain specific information as a \textit{prior}.
	Detailed discussion about prior for haze and rain is provided later in the section.
	We aim to exploit these priors about the weather domain to better adapt the detector for weather affected images.
	To achieve that, we propose a prior-based adversarial training approach using prior estimation network (PEN) and prior adversarial loss (PAL).
	
	Let $\mathcal{P}_{l}$ be PEN module introduced after the $l^{th}$ conv block of $\mathcal{F}$ and let $Z^{src}_{il}$ be the corresponding domain specific prior for any image, $x^s_i \in \mathcal{S}$.
	Then the PAL for the source domain is defined as follows,
	\vskip -17.5 pt
	\begin{align}
	\mathcal{L}^{src}_{pal_{cl}} &= \frac{1}{n_s U V}\sum_{i=1}^{n_s}\sum_{j=1}^{U}\sum_{k=1}^{V} (Z^{src}_{il}-\mathcal{P}_{l}(\mathcal{F}_{l}({x}^{s}_i)))_{jk}^2,
	\end{align}
	where, $U$ and $V$ are height and width of domain specific prior $Z^{src}_{il}$ and output feature $\mathcal{F}_{l}({x}^{s}_i)$. $Z^{src}_{il}$ denotes the source image prior, scaled down from image-level prior to match the scale at $l^{th}$ conv block. Similarly, PAL for the target domain images, $x^t_i \in \mathcal{T}$, with the corresponding prior $Z^{tgt}_{il}$ can be defined as,
	\vskip -17.5 pt
	\begin{align}
	\mathcal{L}^{tgt}_{pal_{cl}} &= \frac{1}{n_t U V}\sum_{i=1}^{n_t}\sum_{j=1}^{U}\sum_{k=1}^{V} (Z^{tgt}_{il}-\mathcal{P}_{l}(\mathcal{F}_{l}({x}^{t}_i)))_{jk}^2,
	\end{align}
	where, we apply PAL after conv4 ($l$=4) and conv5 ($l$=5) block (as shown in Fig. \ref{fig:arch}). Hence, the final source and target adversarial losses can be given as,
	\vskip -17.5 pt
	\begin{align}
	\mathcal{L}^{src}_{pal} &= \frac{1}{2}(\mathcal{L}^{src}_{pal_{c5}} + \mathcal{L}^{src}_{pal_{c4}}),\\
	\mathcal{L}^{tgt}_{pal} &= \frac{1}{2}(\mathcal{L}^{tgt}_{pal_{c5}} + \mathcal{L}^{tgt}_{pal_{c4}}).
	\end{align}
	
	The prior estimation networks ($\mathcal{P}_5$ and $\mathcal{P}_4$) predict the weather-specific prior from the features extracted from $\mathcal{F}$.
	However, the feature extractor network $\mathcal{F}$ is trained to fool the PEN modules by producing features that are weather-invariant (free from weather-specific priors) and prevents the PEN modules from correctly estimating the weather-specific prior.
	Since, this type of training includes prior prediction and is also reminiscent of the adversarial learning used in domain adaptation, we term this loss as prior-adversarial loss.
	At convergence, the feature extractor network $\mathcal{F}$ should have devoid itself from any weather-specific information and as a result both prior estimation networks $\mathcal{P}_5$ and $\mathcal{P}_4$ should not be able to correctly estimate the prior.
	\textit{Note that our goal at convergence is not to estimate the correct prior, but rather to learn weather-invariant features so that the detection network is able to  generalize well to the target domain}.
	This training procedure can be expressed as the following optimization,
	\vskip -9.5 pt
	\begin{equation}\label{eq:adv_objective}
	\underset{\mathcal{F}}{\text{max}} \ \underset{\mathcal{P}}{\text{min}} \ \ \ \mathcal{L}^{src}_{pal} + \mathcal{L}^{tgt}_{pal}.
	\end{equation}
	
	Furthermore, in the conventional domain adaptation, a single label is assigned for entire target image to train the domain discriminator (Fig. \ref{fig:img1})(c)).
	By doing this, it is assumed that the entire image has undergone a constant domain shift.
	However this is not true in the case of weather-affected images, where degradations vary spatially (Fig. \ref{fig:img1})(b)).
	In such cases, the assumption of constant domain shift leads to incorrect alignment especially in the regions of minimal degradations. Incorporating the weather-specific priors overcomes this issue as these priors are spatially varying and are directly correlated with the amount of degradations. Hence, utilizing the weather-specific prior results in better alignment.
	
	\subsubsection{Haze prior}\label{subsubsec:haze_prior}
	The effect of haze on images has been extensively studied in the literature \cite{fattal2008single,he2011single,zhang2018densely,li2017haze,ancuti2018ntire,zhang2018multi,zhang2019joint}.
	Most existing image dehazing methods rely on the atmospheric scattering model for representing image degradations under hazy conditions and is defined as,
	\vskip -12.5 pt
	\begin{equation}
	{I}({z})={J(z)}t({z})+{A(z)}(1-t({z})),
	\label{eq:haze}
	\end{equation}
	where ${I}$ is the observed hazy image, $J$ is the true scene radiance, ${A}$ is the global atmospheric light, indicating the intensity of the ambient light, $t$ is the transmission map and $z$ is the pixel location.
	The transmission map is a distance-dependent factor that affects the fraction of light that reaches the camera sensor.
	When the atmospheric light $A$ is homogeneous, the transmission map can be expressed as $t{(z)}=e^{-\beta d({z})}$, where $\beta$ represents the attenuation coefficient of the atmosphere and $d$ is the scene depth.    
	
	Typically, existing dehazing methods first estimate the transmission map and the atmospheric light, which are then used in Eq. \eqref{eq:haze} to recover the observed radiance or clean image.
	The transmission map contains important information about the haze domain, specifically representing the light attenuation factor.
	We use this transmission as a domain prior for supervising the prior estimation (PEN) while adapting to hazy conditions.
	\textit{Furthermore, instead of depending on the actual ground-truth transmission maps, we use dark channel prior \cite{he2011single} to estimate the transmission maps. Hence, no additional human annotation efforts are required for obtaining the haze prior.}
	
	\subsubsection{Rain prior}\label{subsubsec:rain_prior}
	Similar to dehazing, image deraining methods \cite{Authors16,Authors18,Authors17e,li2019single,you2015adherent} also assume a mathematical model to represent the degradation process and is defined as follows,
	\vskip -15.0 pt
	\begin{equation}
	{I}({z})={J(z)}+{R(z)},
	\label{eq:rain}
	\end{equation}
	where ${I}$ is the observed rainy image, $J$ is the desired clean image, and ${R}$ is the rain residue.
	This formulation models rainy image as a superposition of the clean background image with the rain residue.
	The rain residue contains domain specific information about the rain for a particular image and hence, can be used as a domain specific prior for  supervising the prior estimation network (PEN) while adapting to rainy conditions.
	\textit{Similar to the haze, we do not rely on the actual ground-truth rain residue.
		Instead, we estimate the rain residue using the rain layer prior described in \cite{Authors16} thereby, avoiding the use of expensive human annotation efforts for obtaining the rain prior.}\\
	
	In both cases discussed above (haze prior and rain prior), we do not use any ground-truth labels to estimate respective priors.
	Hence, our overall approach still falls into the category of unsupervised adaptation.
	Furthermore, these priors can be pre-computed for the training images to reduce the computational overhead during the learning process. Additionally, the prior computation is not required during inference and hence, the proposed adaptation method does not result in any computational overhead. 
	
	
	\subsection{Residual Feature Recovery Block}\label{subsec:rrb}
	As discussed earlier, weather-degradations introduce distortions in the feature space. In order to aid the de-distortion process, we introduce a set of residual feature recovery blocks (RFRBs) in the target  feed-forward pipeline.
	This  is inspired from the residual transfer network method proposed in \cite{long2016unsupervised}.
	Let $\Delta\mathcal{F}_l$ be the residual feature recovery block at the $l^{th}$ conv block. 
	The target domain image feedforward is modified to include the residual feature recovery block.
	For $\Delta\mathcal{F}_l$ the feed-forward equation at the $l^{th}$ conv block can be written as,
	\vskip -7.5 pt
	\begin{equation}\label{eq:rfm_equation}
	\mathcal{\hat{F}}_{l}(x^t_i) \ = \ \mathcal{F}_{l}(x^t_i) \ + \ \Delta\mathcal{F}_{l}(\mathcal{F}_{l-1}(x^t_i)), 
	\end{equation}
	where, $\mathcal{F}_{l}(x^t_i)$ indicates the feature extracted from the $l^{th}$ conv block for any image $x^t_i$ sampled from the target domain using the feature extractor network $\mathcal{F}$, $\Delta\mathcal{F}_{l}(\mathcal{F}_{l-1}(x^t_i))$ indicates the residual features extracted from the output ${l-1}^{th}$ conv block, and $\mathcal{\hat{F}}_{l}(x^t_i)$ indicates the feature extracted from the $l^{th}$ conv block for any image $x^t_i \in \mathcal{T}$ with RFRB modified feedforward.
	The RFRB modules are also illustrated in Fig. \ref{fig:arch}, as shown in the target feedforward pipeline.  It has no effect on source feedforward pipeline. In our case, we utilize RFRB at both conv4 ($\Delta\mathcal{F}_{4}$) and conv5 ($\Delta\mathcal{F}_{5}$) blocks. Additionally, the effect of residual feature is regularized by enforcing the norm constraints on the residual features.
	The regularization loss for RFRBs, $\Delta\mathcal{F}_{4}$ and $\Delta\mathcal{F}_{5}$ is defined as,
	\vskip -5.0 pt
	\begin{equation}\label{eq:regularization}
	\mathcal{L}_{reg} = \frac{1}{n_t} \sum_{i=1}^{n_t} \sum_{l=4,5} \|\Delta\mathcal{F}_l(\mathcal{F}_{l-1}(x^t_i))\|_1,
	\end{equation}
	
	\subsection{Overall Loss}\label{subsec:final_training_objective}
	The overall loss for training the network is defined as,
	\vskip -14.0 pt
	\begin{align}\label{eq:final_objective}
	\underset{\mathcal{P}}{\text{max}} \ \underset{\mathcal{F}, \Delta\mathcal{F}, \mathcal{G}}{\text{min}} \ \ \ \mathcal{L}^{src}_{det} &- \mathcal{L}_{adv} + \lambda  \mathcal{L}_{reg}, \;\;\text{where}\\
	\mathcal{L}_{adv} &= \frac{1}{2} (\mathcal{L}^{src}_{pal}+\mathcal{L}^{tgt}_{pal}).
	\end{align}
	
	Here, $\mathcal{F}$ represents the feature extractor network, $\mathcal{P}$ denotes both prior estimation network employed after conv4 and conv5 blocks, i.e., $\mathcal{P}$=$\{\mathcal{P}_5, \mathcal{P}_4\}$, and $\Delta\mathcal{F}$=$\{\Delta\mathcal{F}_4, \Delta\mathcal{F}_5\}$ represents RFRB at both conv4 and conv5 blocks.
	Also, $\mathcal{L}^{src}_{det}$ is the source detection loss, $\mathcal{L}_{reg}$ is the regularization loss, and $\mathcal{L}_{adv}$ is the overall adversarial loss used for prior-based adversarial training.
	
	\section{Experiments and Results}\label{sec:experiments}
	
	\subsection{Implementation details}\label{subsec:implementation_details}
	
	We  follow the training protocol of \cite{Saito2018StrongWeakDA,Chen2018DomainAF} for training the Faster-RCNN network. The backbone network for all experiments is VGG16 network \cite{simonyan2014very}. We model the residuals using RFRB for the convolution blocks C4 and C5 of the VGG16 network. The PA loss is applied to only these conv blocks modeled with RFRBs. The PA loss is designed based on the adaptation setting (Haze or Rain). The parameters of the first two conv blocks are frozen similar to \cite{Saito2018StrongWeakDA,Chen2018DomainAF}. The detailed network architecture for RFRBs, PEN and the discriminator are provided in supplementary material. During training, we set shorter side of the image to 600 with ROI alignment. We train all networks for 70K iterations. For the first 50K iterations, the learning rate is set equal to 0.001 and for the last 20K iterations it is set equal to 0.0001. We report the performance based on the trained model after 70K iterations. We set $\lambda$ equal to 0.1 for all experiments.
	
	In addition to comparison with recent methods, we also perform an ablation study where we evaluate the following  configurations to  analyze the effectiveness of different components in the network. Note that we progressively add additional components which enables us to gauge the performance improvements obtained by each of them,
	\begin{itemize}[topsep=0pt,noitemsep,leftmargin=*]
		\item \textbf{FRCNN:} Source only baseline experiment where Faster-RCNN is trained on the source dataset. 
		\item \textbf{FRCNN+D$_5$:} Domain adaptation baseline experiment consisting of Faster-RCNN with domain discriminator after conv5 supervised by the domain adversarial loss. 
		\item \textbf{FRCNN+D$_5$+R$_5$:} Starting with FRCNN+D$_5$ as the base configuration, we add an RFRB block after conv4 in the  Faster-RCNN. This experiment enables us to understand the contribution of the RFRB block. 
		\item \textbf{FRCNN+P$_5$+R$_5$:} We start with FRCNN+D$_5$+R$_5$ configuration and replace domain discriminator and domain adversarial loss with prior estimation network (PEN) and prior adversarial loss (PAL). With this experiment, we show the importance of training with the proposed prior-adversarial loss. 
		\item \textbf{FRCNN+P$_{45}$+R$_{45}$:} Finally, we perform the prior-based feature alignment at two scales: conv4 and conv5. Starting with  FRCNN+P$_5$+R$_5$ configuration, we  add an  RFRB block after conv3 and a PEN module after conv4. This experiment corresponds to the configuration depicted in Fig. \ref{fig:arch}. This experiment demonstrates the efficacy of the overall method in addition to  establishing the importance of aligning features at multiple levels in the network.
	\end{itemize}
	Following the protocol set by the existing methods \cite{Chen2018DomainAF,Shan2018,Saito2018StrongWeakDA}, we use mean average precision (mAP) scores for performance comparison. 
	
	\subsection{Adaptation to hazy conditions}\label{subsec:adaptation_in_haze}
	In this section, we present the results corresponding to adaptation to hazy conditions on the following  datasets: (i)  Cityscapes $\rightarrow$ Foggy-Cityscapes  \cite{Sakaridis2018SemanticFS}, (ii) Cityscapes $\rightarrow$ RTTS \cite{li2019benchmarking}, and (iii) WIDER  \cite{yang2016wider} $\rightarrow$  UFDD-Haze \cite{nada2018pushing}. In the first two  experiments, we consider Cityscapes \cite{cordts2016cityscapes} as the source domain. Note that the Cityscapes dataset contains images captured in clear weather conditions. \\
	
	\noindent \textbf{Cityscapes $\rightarrow$ Foggy-Cityscapes:} In this experiment, we adapt from Cityscapes to  Foggy-Cityscapes \cite{Sakaridis2018SemanticFS}.  The Foggy-Cityscapes dataset was recently proposed in \cite{Sakaridis2018SemanticFS} to study the detection algorithms in the case of hazy weather conditions. Foggy-Cityscapes is derived from Cityscapes dataset by simulating fog on the clear weather images of Cityscapes. Both Cityscapes and Foggy-Cityscapes have the same number of categories which include, car, truck, motorcycle/bike, train, bus, rider and person. Similar to \cite{Chen2018DomainAF}, \cite{Saito2018StrongWeakDA},  we utilize 2975 images of both Cityscapes and Foggy-Cityscapes for training. Note that we use  annotations only from the source dataset (Cityscapes) for training the detection pipeline. For evaluation we consider a non overlapping validation set of 500 images provided by the Foggy-Cityscapes dataset. 
	
	We compare the proposed method with two categories of approaches: \textit{(i) Dehaze+Detect:}  Here, we employ dehazing network as pre-processing step and perform detection using Faster-RCNN trained on source (clean) images. For pre-processing, we chose two recent dehazing algorithms:  DCPDN \cite{zhang2018densely} and Grid-Dehaze \cite{liu2019griddehazenet}.  \textit{(i) DA-based methods:} Here, we compare with following recent domain-adaptive detection approaches: DA-Faster \cite{Chen2018DomainAF}, SWDA \cite{Saito2018StrongWeakDA}, DiversifyMatch \cite{kim2019diversify}, Mean Teacher with Object Relations (MTOR) \cite{cai2019exploring}, Selective Cross-Domain Alignment (SCDA) \cite{zhu2019adapting} and Noisy Labeling \cite{khodabandeh2019robust}. The corresponding results are presented in Table \ref{tab:c2f}. 
	
	%
	\begin{figure}[t!]
		\begin{center}
			\includegraphics[width=1\linewidth]{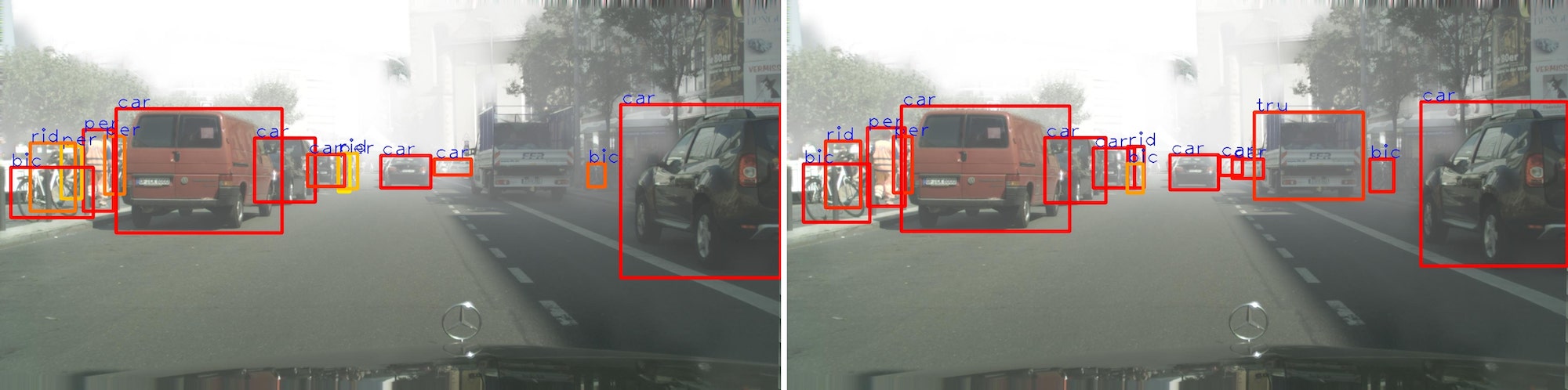}
			\includegraphics[width=.4\linewidth]{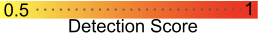}\\
			\vskip -5pt 
			(a) \hskip 130pt (b) 
		\end{center}
		
		\vskip -20pt \caption{Detection results on Foggy-Cityscapes. (a) DA-Faster RCNN \cite{Chen2018DomainAF}. (b) Proposed method. The bounding boxes are colored based on the detector confidence using the color map as shown.	DA-Faster-RCNN produces the detections with low confidence in addition to missing the truck class in both samples. In contrast, the proposed  method is able to output high confidence detections without missing any objects. }
		\label{fig:results_foggy_cityscapes}
	\end{figure}
	
	\begin{table}[t!]
		
		\caption{Performance comparison for the Cityscapes $\rightarrow$ Foggy-Cityscapes experiment. }
		\label{tab:c2f}
		\vskip -9pt
		\resizebox{.81\linewidth}{!}{
			\begin{tabular}{|l|l|cccccccc|c|}
				\hline
				\multicolumn{2}{|l|}{Method}                                                 & prsn & rider & car  & truc & bus  & train & bike & bcycle & mAP \\ \hline \hline
				\multirow{1}{*}{Baseline}                  & FRCNN \cite{ren2015faster}                            & 25.8   & 33.7  & 35.2 & 13.0  & 28.2 & 9.1   & 18.7 & 31.4  & 24.4    \\
				\hline\hline
				\multirow{2}{*}{Dehaze}      & DCPDN \cite{zhang2018densely}                           & 27.9   & 36.2  & 35.2 & 16.0  & 28.3 & 10.2   & 24.6 & 32.5  & 26.4    \\
				& Grid-Dehaze \cite{liu2019griddehazenet} 					&  {29.7}  & {40.4} & 40.3  & {21.3} & 30.0  & {9.1}  & {25.6} & 36.7 & 29.2   \\ \hline\hline
				\multirow{6}{*}{DA-Methods}                
				& DAFaster \cite{Chen2018DomainAF}                          & 25.0   & 31.0  & 40.5 & 22.1  & 35.3 & 20.2  & 20.0 & 27.1  & 27.6     \\ 
				& SCDA \cite{zhu2019adapting}                   & 33.5   & 38.0  & 48.5 & 26.5  & 39.0 & 23.3  & 28.0 & 33.6  & 33.8   \\ 
				& SWDA \cite{Saito2018StrongWeakDA}                 & 29.9   & 42.3  & 43.5 & 24.5  & 36.2 & 32.6  & 30.0 & 35.3  & 34.3      \\ 
				& DM \cite{kim2019diversify} & 30.8   & 40.5  & 44.3 & \bf{27.2}  & 38.4 & 34.5  & 28.4 & 32.2  & 34.6    \\ 
				& MTOR \cite{cai2019exploring}   & 30.6   & 41.4  & 44.0 & 21.9  & 38.6 & \bf{40.6}  & 28.3 & 35.6   & 35.1    \\
				& NL \cite{khodabandeh2019robust} & 35.1   & 42.1  & 49.2 & 30.1  & 45.3 & 26.9  & 26.8 & 36.0  & 36.5        \\ \hline\hline
				\multirow{4}{*}{Ours}                 
				& FRCNN+D$_5$                          & 30.9   & 38.5  & 44.0 & 19.6  & 32.9 & 17.9  & 24.1 & 32.4  & 30.0    \\ 
				& FRCNN+D$_5$+R$_5$                    & 32.8   & 44.7  & 49.9 & 22.3  & 31.7 & 17.3  & 26.9 & 37.5  & 32.9    \\ 
				& FRCNN+P$_5$+R$_5$                    & 33.4   & 42.8  & 50.0 & 24.2  & 40.8 & 30.4  & 33.1 & 37.5  & 36.5    \\ 
				& FRCNN+P$_{45}$+R$_{45}$& \bf{36.4}   & \bf{47.3}  & \bf{51.7} & 22.8  & \bf{47.6} & 34.1  & \bf{36.0} & \bf{38.7}  & \bf{39.3}    \\ \hline
			\end{tabular}
		}
	\end{table}

	It can be observed from Table \ref{tab:c2f}, that the performance of source-only training of Faster-RCNN is in general poor in the hazy conditions. Adding DCPDN and Gird-Dehaze as preprocessing step improves the performance by $\sim$2\% and $\sim$4\%, respectively. Compared to the domain-adaptive detection approaches,   pre-processing + detection results in lower performance gains. This is because even after applying dehazing there still remains some domain shift as discussed in Sec. \ref{sec:introduction}. Hence, using adaptation would be a better approach for mitigating the domain shift. Here, the use of simple domain adaptation \cite{ganin2014unsupervised} (FRCNN+D$_5$) improves the source-only performance. The addition of RFRB$_5$ (FRCNN+D$_5$+R$_5$) results in further improvements, thus indicating the importance of RFRB blocks. However, the conventional domain adaptation loss assumes constant domain shift across the entire image, resulting in incorrect alignment. The use of prior-adversarial loss  (FRCNN+P$_5$+R$_5$) overcomes this issue. We achieved 3.6\% improvement in overall mAP scores, thus demonstrating the effectiveness of the proposed prior-adversarial training. Note that, FRCNN+P$_5$+R$_5$ baseline achieves comparable performance with state-of-the-art. Finally, by performing prior-adversarial adaptation at an additional scale (FRCNN+P$_{45}$+R$_{45}$), we achieve further improvements which surpasses the existing best approach \cite{khodabandeh2019robust} by 2.8\%. Fig. \ref{fig:results_foggy_cityscapes} shows sample qualitative detection results corresponding to the images from Foggy-Cityscapes. Results for the proposed method are compared with DA-Faster-RCNN \cite{Chen2018DomainAF}. It can be observed that the proposed method  is able to generate comparatively high quality detections.  
	
	We summarize our observations as follows: (i) Using dehazing as a pre-processing step results in minimal improvements over the baseline Faster-RCNN. Domain adaptive approaches perform better in general. (ii) The proposed method outperforms other methods in the overall scores while achieving the best performance in most of the classes. See supplementary material for more ablations. \\
	
	\noindent \textbf{Cityscapes $\rightarrow$ RTTS:} In this experiment,  we adapt from Cityscapes to the RTTS dataset \cite{li2019benchmarking} . RTTS is a subset of a larger RESIDE dataset \cite{li2019benchmarking}, and it contains 4,807 unannotated and 4,322 annotated real-world hazy images covering mostly traffic and driving scenarios. We use the unannotated 4,807 images for training the domain adaptation process. The evaluation is performed on the annotated 4,322 images. RTTS has total five categories, namely motorcycle/bike, person, bicycle, bus and car. This dataset is the largest available dataset for object detection under real world hazy conditions.  
	
	In Table \ref{table:city2rtts}, the results of the proposed method are compared with Faster-RCNN \cite{ren2015faster}, DA-Faster \cite{Chen2018DomainAF} and SWDA \cite{Saito2018StrongWeakDA} and the dehaze+detection baseline as well. For RTTS dataset, the pre-processing with DCPDN improves the Faster-RCNN performance by $\sim$1\%. Surprisingly, Grid-Dehaze does not help the Faster-RCNN baseline and results in even worse performance. Whereas, the proposed method achieves an improvement of 3.1\% over the baseline Faster-RCNN (source-only training), while outperforming the other recent methods.\\
	
	\begin{table}[t!]
		\centering
		\caption{Performance comparison for the Cityscapes $\rightarrow$ RTTS experiment.}
		\label{table:city2rtts}
		\vskip -8pt
		\resizebox{.65\linewidth}{!}{
			\begin{tabular}{|l|l|ccccc|c|}
				\hline 
				
				\multicolumn{2}{|l|}{Method} & prsn & car  & bus  & bike & bcycle & mAP  \\ \hline\hline
				\multirow{1}{*}{Baseline}&FRCNN  \cite{ren2015faster}                 & 46.6 & 39.8 & 11.7 & 19.0 & 37.0   & 30.9 \\ \hline\hline
				\multirow{2}{*}{Dehaze}&DCPDN \cite{zhang2018densely}                 & \textbf{48.7} & 39.5 & 12.9 & 19.7 & 37.5   & 31.6 \\ 
				& Grid-Dehaze \cite{liu2019griddehazenet}    & 29.7 & 25.4 & 10.9 & 13.0 & 21.4   & 20.0 \\ \hline\hline
				\multirow{3}{*}{DA}
				&DAFaster \cite{Chen2018DomainAF}   & 37.7 & 48.0 & 14.0 & \textbf{27.9} & 36.0   & 32.8 \\ 
				&SWDA  \cite{Saito2018StrongWeakDA} & 42.0 & 46.9  & 15.8 & 25.3 & 37.8 & 33.5 \\ \hline\hline
				Ours&Proposed                & 37.4 & \textbf{54.7} & \textbf{17.2} & 22.5 & \textbf{38.5}   & \textbf{34.1} \\ \hline
			\end{tabular}
		}
	\end{table}
	
	\begin{table}[b!]
		\centering
		\caption{Results (mAP) of the adaptation experiments from WIDER-Face to UFDD Haze and Rain.}
		\label{tab:wider}
		\vskip -8pt 
		\resizebox{0.35 \linewidth}{!}{
			\begin{tabular}{|l|cc|}
				\hline
				Method          & UFDD-Haze & UFDD-Rain \\ \hline\hline
				FRCNN \cite{ren2015faster}     & 46.4     & 54.8     \\
				DAFaster \cite{Chen2018DomainAF} & 52.1     & 58.2     \\ 
				SWDA \cite{Saito2018StrongWeakDA}  & 55.5    & 60.0    \\ \hline \hline
				Proposed  & \textbf{58.5}     & \textbf{62.1}     \\ \hline
			\end{tabular}
		}
	\end{table}

	\noindent \textbf{WIDER-Face $\rightarrow$ UFDD-Haze:} Recently, Nada \etal \cite{nada2018pushing} published a benchmark face detection dataset which consists of real-world images captured under different weather-based conditions such as haze and rain. Specifically, this dataset consists of  442 images under the haze category. Since, face detection is closely related to the task of object detection, we evaluate our framework by adapting from WIDER-Face \cite{yang2016wider} dataset to UFDD-Haze dataset. WIDER-Face is a large-scale face detection dataset with approximately 32,000 images and 199K face annotations. The results corresponding to this adaptation experiment are shown in Table \ref{tab:wider}. It can be observed from this table that the proposed method achieves better performance as compared to the other methods. \\

	\subsection{Adaptation to rainy conditions}\label{subsec:adaptation_in_rain}
	
	\begin{table}[b!]
		\caption{Performance comparison for the Cityscapes $\rightarrow$ Rainy-Cityscapes experiment. }
		\label{tab:city2rainy}
		\vskip -9pt
		\resizebox{.75\linewidth}{!}{
			\begin{tabular}{|l|l|cccccccc|c|}
				\hline
				\multicolumn{2}{|l|}{Method}                                                 & prsn & rider & car  & truc & bus  & train & bike & bcycle & mAP \\ \hline \hline
				\multirow{1}{*}{Baseline}                  & FRCNN                               & 21.6          & 19.5          & 38.0          & 12.6          & 30.1          & 24.1          & 12.9          & 15.4          & 21.8       \\ \hline\hline
				\multirow{2}{*}{Derain} & DDN \cite{fu2017removing}           &  27.1         & 30.3         & 50.7          & 23.1          & 39.4          & 18.5           & 21.2         & 24.0         & 29.3    \\  
				& SPANet \cite{wang2019spatial}          &  24.9         & 28.9         & 48.1          & 21.4          & 34.8          & 16.8          & 17.6          & 20.8        & 26.7     \\\hline  \hline
				\multirow{2}{*}{DA} & DAFaster \cite{Chen2018DomainAF}           &  26.9          & 28.1          & 50.6          & 23.2          & 39.3          & 4.7           & 17.1          & 20.2          & 26.3     \\  
				& SWDA \cite{Saito2018StrongWeakDA}          & 29.6  &   \textbf{38.0} &   52.1&   27.9 & \textbf{49.8}  &    28.7&  24.1& 25.4  &     34.5 \\\hline  \hline
				\multirow{5}{*}{Ours}                  
				& FRCNN+D$_{5}$                          &   29.1  &34.8  & 52.0  & 22.0 & 41.8  & 20.4  & 18.1& 23.3  &  30.2    \\ 
				& FRCNN+D$_5$+R$_5$                    &   28.8  &  33.1 &   51.7&   22.3 & 41.8 & 24.9   & 22.2  & 24.6   &  31.2   \\ 
				& FRCNN+P$_5$+R$_5$                    & 29.7    & 34.3  & 52.5 & 23.6   &  47.9 &  32.5 & 24.0  &  25.5  &  33.8   \\ 
				& FRCNN+P$_{45}$+R$_{45}$                  &\textbf{31.3} & {34.8} & \textbf{57.8} & \textbf{29.3} & {48.6} & \textbf{34.4} & \textbf{25.4} & \textbf{27.3} & \textbf{36.1}     \\ \hline
			\end{tabular}
		}
	\end{table}

	In this section, we present the results of adaptation to rainy conditions. Due to lack of appropriate datasets for this particular setting,  we create a new rainy dataset called Rainy-Cityscapes and it is derived from Cityscapes. It has the same number of images for training and validation as Foggy-Cityscapes. First, we discuss the  simulation process used to create the dataset, followed by a discussion of the evaluation and comparison of the proposed method with other methods.\\
	
	\noindent \textbf{Rainy-Cityscapes: } Similar to Foggy-Cityscapes, we use a subset of 3475 images from Cityscapes to create synthetic rain dataset. Using \cite{synth-rain},  several  masks containing artificial rain streaks are synthesized. The rain streaks are created using different Gaussian noise levels and multiple rotation angles between $70^\circ$ and  $110^\circ$. Next, for every image in the subset of the Cityscapes dataset, we pick a random rain mask and blend it onto the image to generate the synthetic rainy image. More details and example images are provided in supplementary material. \\

	\noindent \textbf{Cityscapes $\rightarrow$ Rainy-Cityscapes:} In this  experiment, we adapt from  Cityscapes to Rainy-Cityscapes. We compare the proposed method with recent methods such as DA-Faster \cite{Chen2018DomainAF} and SWDA \cite{Saito2018StrongWeakDA}. Additionally, we also evaluate performance of two derain+detect baselines, where state of the art methods such as DDN \cite{fu2017removing} and SPANet \cite{wang2019spatial} are used as a pre-processing step to the Faster-RCNN trained on source (clean) images. From the Table \ref{tab:city2rainy} we  observe that such methods provide reasonable improvements over the Faster-RCNN baseline. However, the performance gains are much lesser as compared to adaptation methods, for the reasons discussed in the earlier sections (Sec. \ref{sec:introduction}, Sec. \ref{subsec:adaptation_in_haze}). Also, it can be observed from Table \ref{tab:city2rainy}, that the proposed method outperforms the other methods by a significant margin. Additionally, we present the results of the ablation study consisting of the experiments listed in Sec. \ref{subsec:implementation_details}.
	The introduction of domain adaptation loss significantly improves the source only Faster-RCNN baseline, resulting in approximately 9\% improvement for FRCNN+D$_5$ baseline in Table \ref{tab:city2rainy}. This performance is further improved by 1\% with the help of residual feature recovery blocks as shown in FRCNN+D$_5$+R$_5$ baseline. When domain adversarial training is replaced with prior adversarial training with PAL, i.e. FRCNN+P$_5$+R$_5$ baseline, we observe 2.5\% improvements, showing effectiveness of the proposed training methodology. Finally, by performing prior adversarial training at multiple scales, the proposed method FRCNN+P$_{45}$+R$_{45}$ observes approximately 2\% improvements and also outperforms the next best method SWDA \cite{Saito2018StrongWeakDA} by 1.6\%. Fig. \ref{fig:rn} illustrates sample detection results obtained using the proposed method as compared to a recent method \cite{Chen2018DomainAF}. The proposed method achieves superior quality detections. \\

	\begin{figure}[t!]
		\begin{center} 
			\includegraphics[width=1\linewidth]{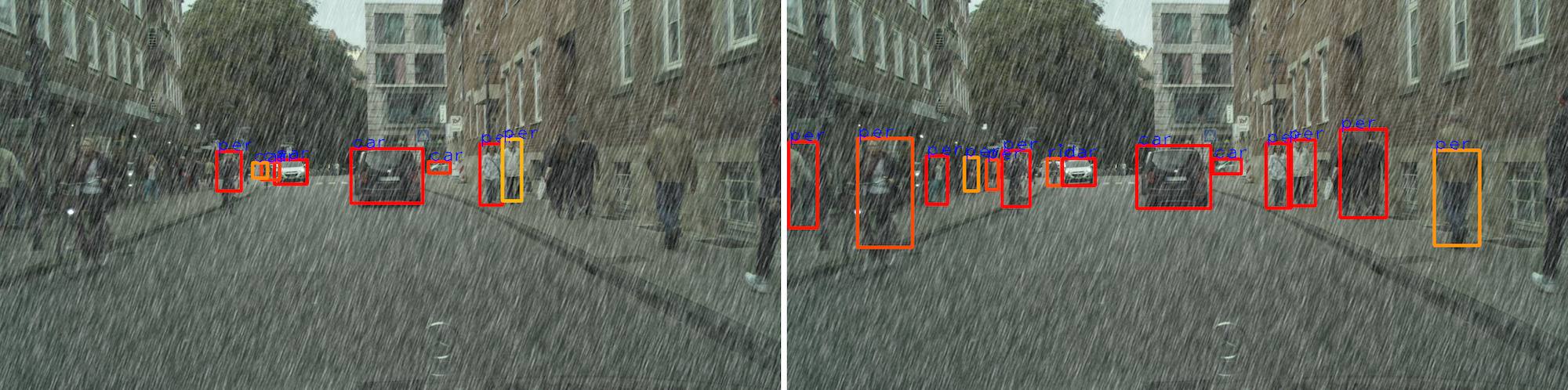}
			\includegraphics[width=.4\linewidth]{figures/map}\\
			\vskip -5pt 
			(a) \hskip 130pt (b) 
		\end{center}
		\vskip -18pt \caption{Detection results on Rainy-Cityscapes. (a) DA-Faster RCNN \cite{Chen2018DomainAF}. (b) Proposed method. The bounding boxes are colored based on the detector confidence using the color map as shown. DA-Faster-RCNN misses several objects in both the samples. In contrast, the proposed  method is able to output high confidence detections without missing any objects. }
		\label{fig:rn}
	\end{figure}

	\noindent \textbf{WIDER-Face $\rightarrow$ UFDD-Rain:} In this experiment, we adapt from WIDER-Face to UFDD-Rain \cite{nada2018pushing}. The UFDD-Rain dataset consists of 628 images collected under  rainy conditions. The results of the proposed method as compared to the other methods are shown in Table \ref{tab:wider}. It can be observed that the proposed method outperforms the source only training by 7.3\%.
	We provide additional details about the proposed method including results and analysis  in the supplementary material.
	
	\section{Conclusions}
	We addressed the problem of adapting object detectors to   hazy and rainy conditions. Based on the observation that these weather conditions cause degradations that can be mathematically modeled and  cause spatially varying distortions in the feature space, we propose a novel prior-adversarial loss that aims at producing weather-invariant features.   Additionally, a set of residual feature recovery blocks are introduced  to learn residual features that can aid efficiently aid the adaptation process.The proposed framework is evaluated on several benchmark datasets such as Foggy-Cityscapes, RTTS and UFDD. Through extensive experiments, we show that our method achieves significant gains over the recent methods in all the datasets. 
	
	\section*{Acknowledgement}
	This work was supported by the NSF grant 1910141
\bibliographystyle{splncs04}
\bibliography{egbib}
\clearpage
\section{Supplementary Material}

\subsection{Additional Results}

\noindent\textbf{Results with ResNet-152}

Table~\ref{tab:resnet_c2f} shows the additional results on the Cityscapes $\rightarrow$ Foggy-Cityscapes experiments, when ResNet-152 network architecture is used as backbone of detection network. From the results we can see that ResNet-152 performs better compared to the corresponding VGG16 baselines. For FRCNN+P$_{45}$+R$_{45}$ baseline with ResNet-152, residual feature recovery blocks and prior estimation networks are applied on fourth and fifth conv block of the network. The results in Table~\ref{tab:resnet_c2f} show that the proposed approach generalizes well for different network architectures.

\begin{table}[t!]
	\caption{Performance comparison for the Cityscapes $\rightarrow$ Foggy-Cityscapes experiment. Red and Blue color fonts show best and second best performance.}
	\label{tab:resnet_c2f}
	\vskip -9pt
	\resizebox{0.7\linewidth}{!}{
		\begin{tabular}{|l|l|cccccccc|c|}
			\hline
			\multicolumn{2}{|l|}{Method}                                                 & prsn & rider & car  & truc & bus  & train & bike & bcycle & mAP \\ \hline \hline
			\multicolumn{2}{|l|}{DAFaster} & 25.0   & 31.0  & 40.5 & 22.1  & 35.3 & 20.2  & 20.0 & 27.1  & 27.6    \\  
			\multicolumn{2}{|l|}{SCDA}     & 33.5   & 38.0  & 48.5 & 26.5  & 39.0 & 23.3  & 28.0 & 33.6  & 33.8    \\ 
			\multicolumn{2}{|l|}{SWDA}     & 29.9   & 42.3  & 43.5 & 24.5  & 36.2 & 32.6  & 30.0 & 35.3  & 34.3    \\  
			\multicolumn{2}{|l|}{DM}       & 30.8   & 40.5  & 44.3 & 27.2  & 38.4 & 34.5  & 28.4 & 32.2  & 34.6    \\  
			\multicolumn{2}{|l|}{MTOR}     & 30.6   & 41.4  & 44.0 & 21.9  & 38.6 & 40.6  & 28.3 & 35.6  & 35.1    \\ 
			\multicolumn{2}{|l|}{NL}       & 35.1   & 42.1  & 49.2 & \bf{\textcolor{red}{30.1}}  & 45.3 & 26.9  & 26.8 & 36.0  & 36.5    \\ \hline \hline
			\multirow{2}{*}{Ours (VGG16)}  & FRCNN  & 25.8  & 33.7 & 35.2  & 13.0 & 28.2  & 9.1  & 18.7  & 31.4  & 24.4\\
			& FRCNN+P$_{45}$+R$_{45}$& \bf{\textcolor{red}{36.4}}   & \bf{\textcolor{red}{47.3}}  & \bf{\textcolor{red}{51.7}} & 22.8  & \bf{\textcolor{red}{47.6}} & 34.1  & \bf{\textcolor{red}{36.0}} & \bf{\textcolor{red}{38.7}}  & \bf{\textcolor{blue}{39.3}}    \\ \hline \hline
			\multirow{2}{*}{Ours (ResNet-152)}  & FRCNN  & 32.4  & 42.2  & 36.0  & 19.8  & 26.4  & 4.7  & 22.7  & 32.6  & 27.1\\
			& FRCNN+P$_{45}$+R$_{45}$& \bf{\textcolor{blue}{34.9}} & \bf{\textcolor{blue}{46.4}}  & \bf{\textcolor{blue}{51.4}} & \bf{\textcolor{blue}{29.2}}  & \bf{\textcolor{blue}{46.3}} & \bf{\textcolor{red}{43.2}}  & \bf{\textcolor{blue}{31.7}} & \bf{\textcolor{blue}{37.0}}  & \bf{\textcolor{red}{40.0}}    \\ \hline
		\end{tabular}
	}
\end{table}

\noindent\textbf{Ablation Analysis}

The Table~\ref{tab:ablation} provides  additional ablation experiments with different network configuration. The analysis is done with VGG-16 network architecture as backbone for detection network.

\begin{table}[t!]
	\centering
	\caption{Results of the ablation experiments from Cityscapes $\rightarrow$ Foggy-Cityscapes. Here, $^*$ indicates additional experiments that are not included in the paper.}
	\label{tab:ablation}
	\vskip -8pt 
	\resizebox{0.2 \linewidth}{!}{
		\begin{tabular}{|l|c|}
			\hline
			Method                   & mAP  \\ \hline \hline
			FRCNN                    & 24.4 \\ \hline
			FRCNN+D$_5$              & 30.0 \\ 
			FRCNN+D$_5$+R$_5$        & 32.9 \\ 
			FRCNN+D$_{45}^*$           & 33.2 \\ \hline
			FRCNN+P$_5$+R$_5$        & 36.5 \\ 
			FRCNN+P$_{45}^*$           & 37.4 \\ 
			FRCNN+P$_{45}$+R$_{45}$  & \textbf{39.3} \\ \hline
		\end{tabular}
	}
\end{table}

\noindent\textbf{Parameter Sensitivity}

\begin{figure}[t!]
	\begin{center}
		\includegraphics[width=.3\linewidth]{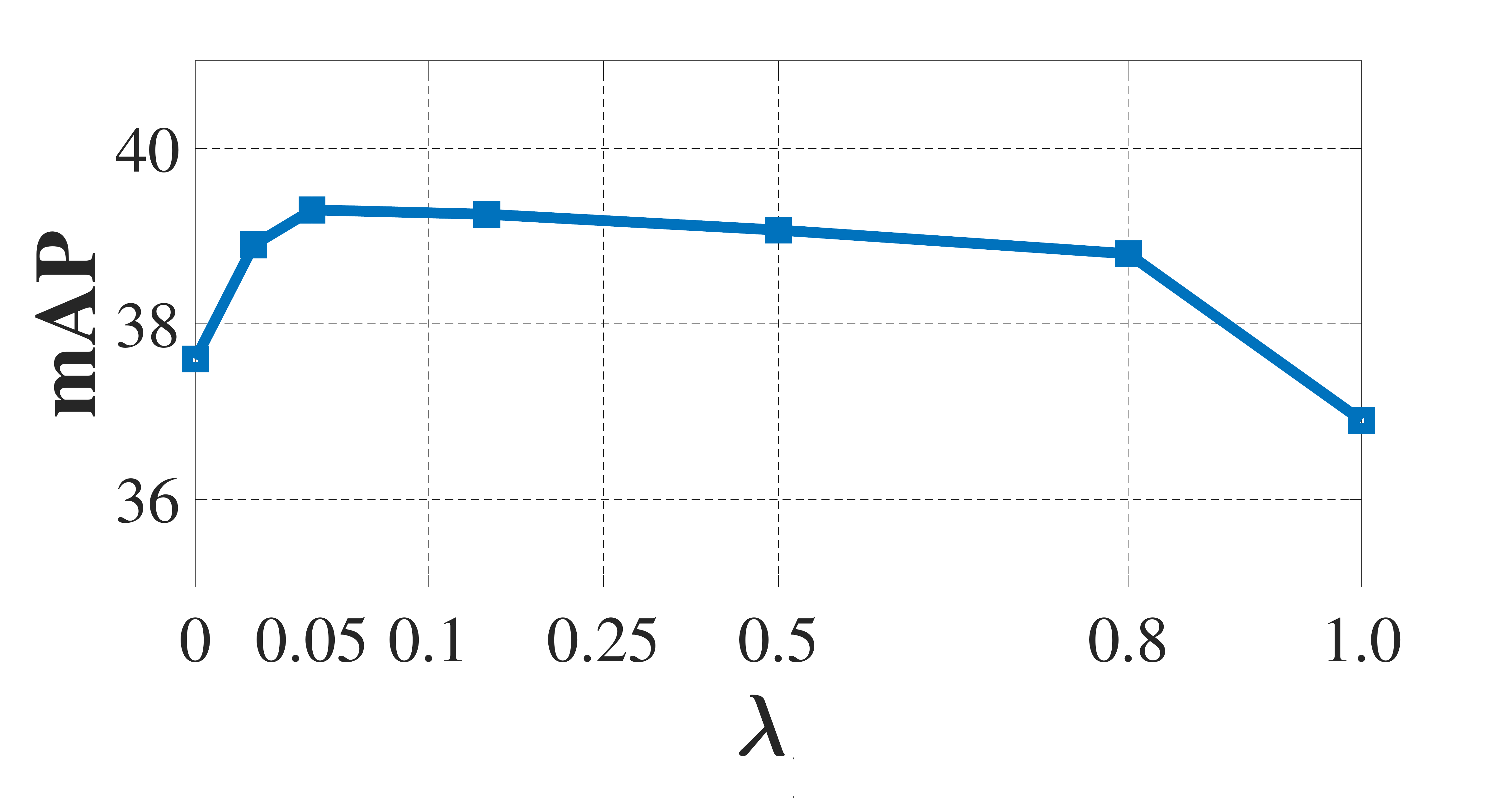}
	\end{center}
	\vskip -25.0pt \caption{Performance sensitivity of proposed approach with varying $\lambda$ parameter.}
	\label{fig:lambda}
\end{figure}

In Fig.\ref{fig:lambda}, we provide sensitivity of the proposed approach with respect to $\lambda$ parameter. The parameter $\lambda$ controls the effect of regularization applied on residual feature norm coming from residual feature recovery blocks. The parameter sensitivity experiment was performed for adaptation from Cityscapes $\rightarrow$ Foggy-Cityscapes with VGG16 network architecture as backbone of detection network.

\subsection{Network configurations}

The network configuration details of different modules such as Residual Feature Recovery Blocks  and Prior Estimation Network  are shown in Table \ref{tab:net2}. 
\begin{table}[t !]
	\centering
	\caption{Network configuration details for Prior Estimation Networks.}
	\vskip -5.0pt
	\label{tab:net1}
	\resizebox{0.3\linewidth}{!}{
		\begin{tabular}{|c|}
			\hline
			\textbf{Prior Estimation Network}   \\ \hline
			Gradient Reversal Layer             \\ \hline
			Conv, 1 $\times$ 1, 64, stride 1, BN, ReLU \\ \hline
			Conv, 3 $\times$ 3, 64, stride 1, BN, ReLU \\ \hline
			Conv, 3 $\times$ 3, 64, stride 1, BN, ReLU \\ \hline
			Conv, 3 $\times$ 3, 3, stride 1, Tanh      \\ \hline
		\end{tabular}
	}
\end{table}

\begin{table}[h!]
	\centering
	\caption{Network configuration details for Residual Feature Recovery Blocks.}
	\vskip -10.0pt
	\label{tab:net2}
	\resizebox{0.7\linewidth}{!}{
		\begin{tabular}{|c|lc|c|}
			\cline{1-1} \cline{4-4}
			\textbf{Residual Feature Recovery Block - Conv4} &  &  & \textbf{Residual Feature Recovery Block - Conv5} \\ \cline{1-1} \cline{4-4} 
			Maxpool, 2 $\times$ 2, stride 2                          &  &  & Maxpool, 2 $\times$ 2, stride 2                          \\ \cline{1-1} \cline{4-4} 
			Conv, 3 $\times$ 3, 256, stride 1, padding 1, ReLU       &  &  & Conv, 3 $\times$ 3, 512, stride 1, padding 1, ReLU       \\ \cline{1-1} \cline{4-4} 
			Conv, 3 $\times$ 3, 512, stride 1, padding 1, ReLU       &  &  & Conv, 3 $\times$ 3, 512, stride 1, padding 1, ReLU       \\ \cline{1-1} \cline{4-4} 
			Conv, 3 $\times$ 3, 512, stride 1, padding 1             &  &  & Conv, 3 $\times$ 3, 512, stride 1, padding 1             \\ \cline{1-1} \cline{4-4} 
		\end{tabular}
	}
\end{table}


\pagebreak

\noindent\textbf{t-SNE Feature Visualization} 

\begin{figure}[ht!]
	\begin{center}
		\includegraphics[width=1.00\linewidth]{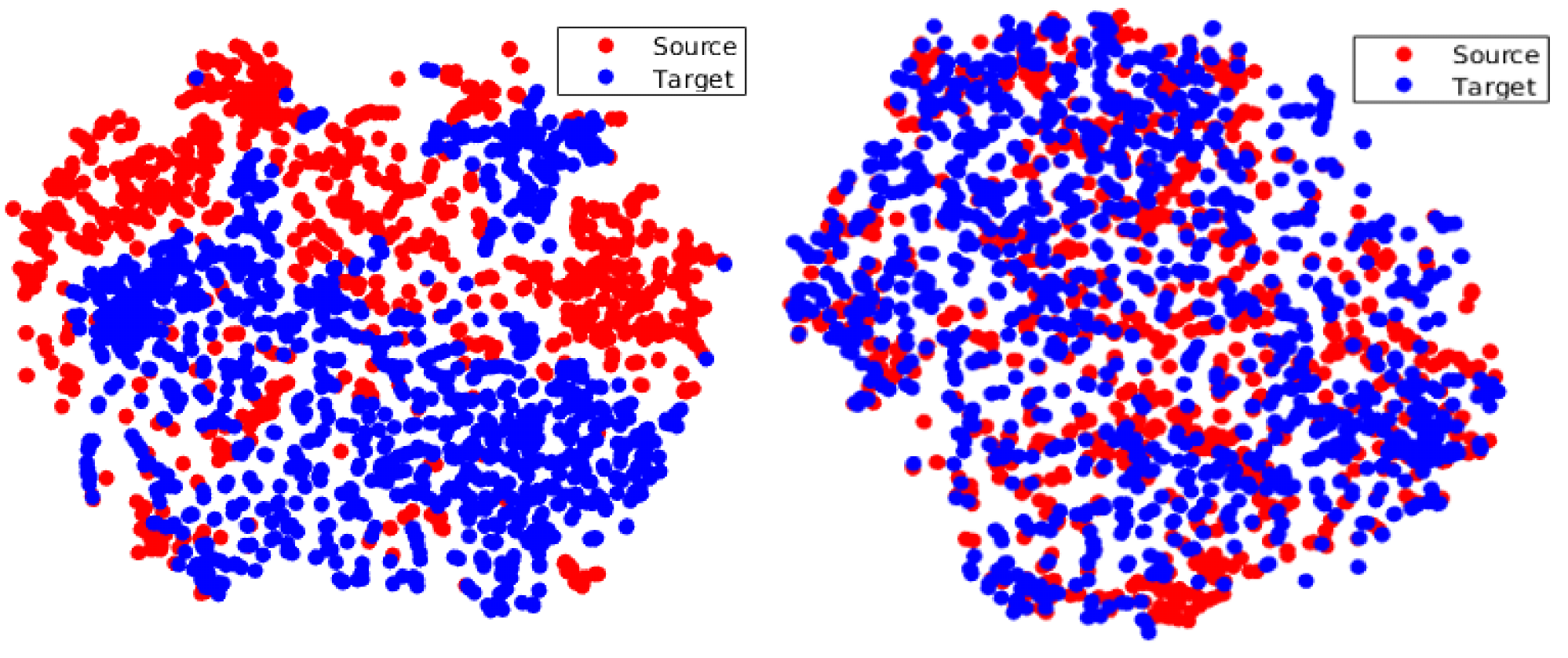}\\
		(a)\hskip160pt(b)
	\end{center}
	\vskip -18pt \caption{Visualization of features using t-SNE plots of different models for Foggy-Cityscapes. (a) Model trained using only the domain adaptive loss. (b) Model trained using the prior adversarial loss. With the domain adaptive loss, the features are not perfectly aligned. Introducing the prior adversarial loss results in better alignment.}
	\label{fig:tsne_fog}
\end{figure}

\pagebreak

\subsection{Qualitative Results}
\noindent\textbf{Cityscaps $\rightarrow$ Foggy-Cityscapes} 

\begin{figure*}[h!]
	\begin{center}
		\includegraphics[width=0.95\linewidth]{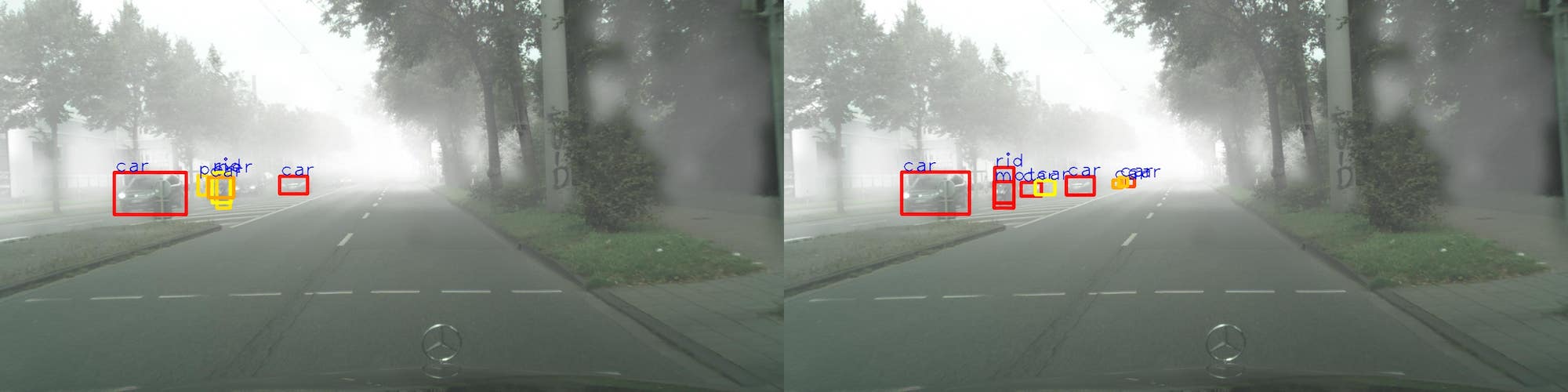}
		\includegraphics[width=0.95\linewidth]{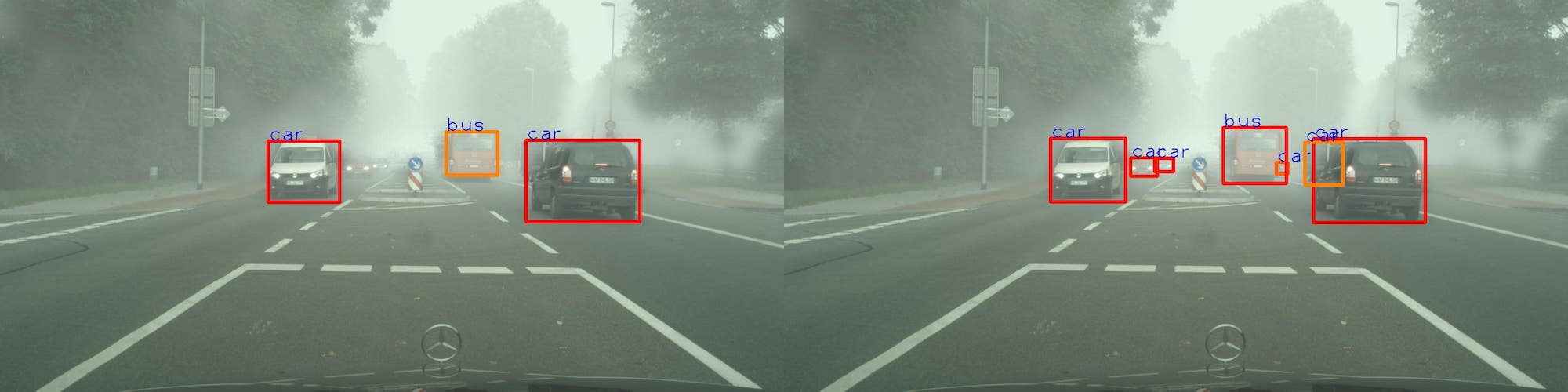}
		\includegraphics[width=0.95\linewidth]{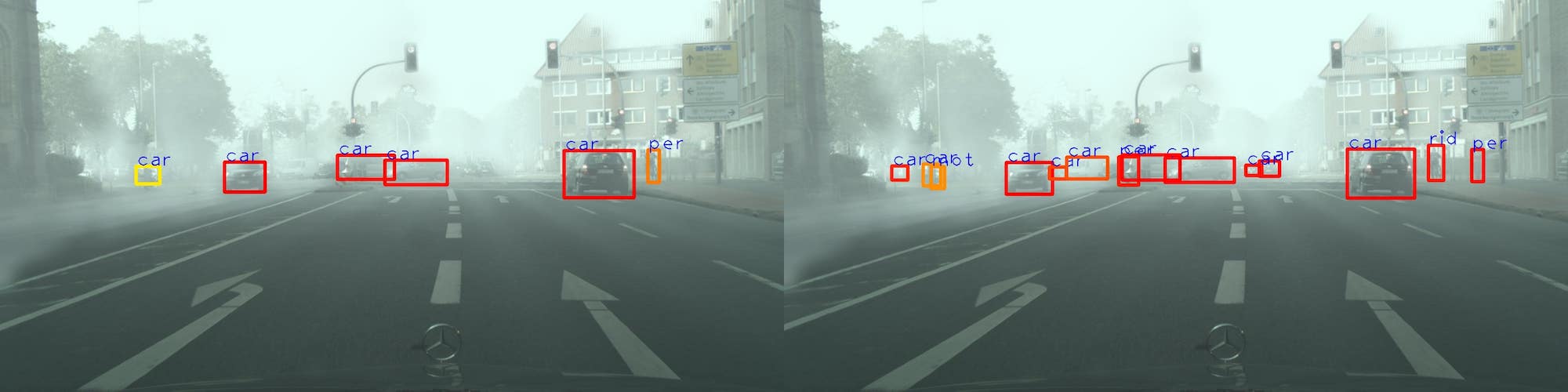}
		\includegraphics[width=0.95\linewidth]{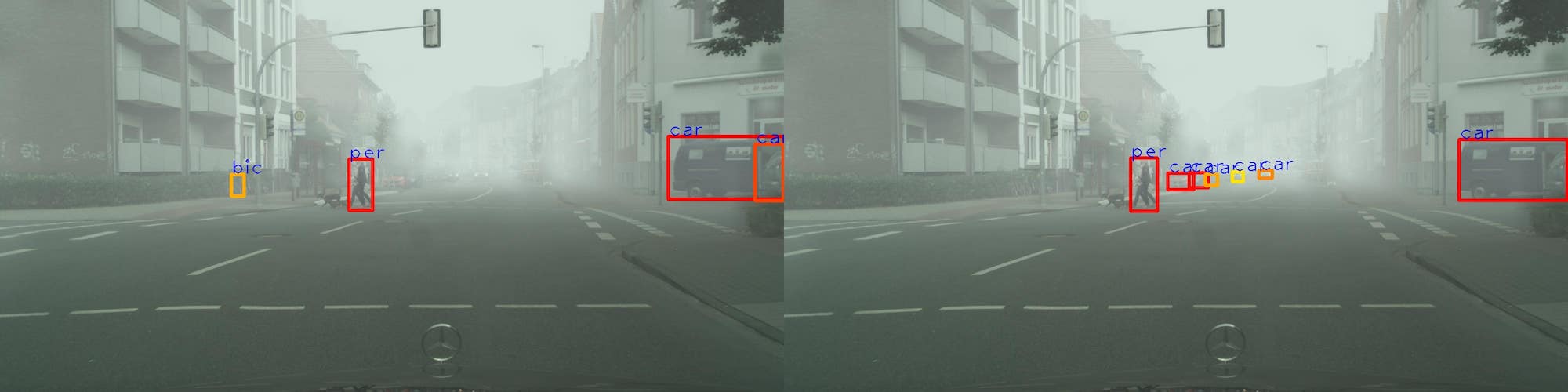}
		\includegraphics[width=0.95\linewidth]{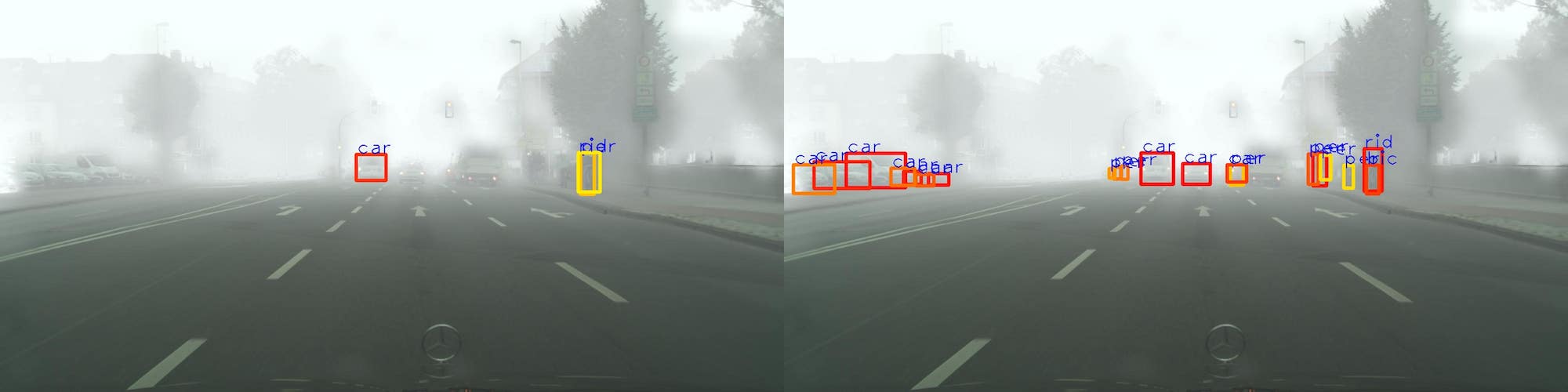}\\
		\includegraphics[width=.25\linewidth]{figures/map}\\
		\vskip -10pt
		(a) \hskip 180pt (b)
	\end{center}
	\label{fig:fg}
	\vskip -15pt \caption{Detection results on Foggy-Cityscapes. (a) DA-Faster RCNN \cite{Chen2018DomainAF} (b) Proposed method. The bounding boxes are colored based on the detector confidence using the color map as shown. As we can see from the figures, the proposed method is able to produce high confidence predictions and is able to detect more objects in the image.}
\end{figure*}

\pagebreak
\noindent\textbf{Cityscaps $\rightarrow$ Rainy-Cityscapes}\\
\begin{figure*}[h!]
	\begin{center}
		\includegraphics[width=0.82\linewidth]{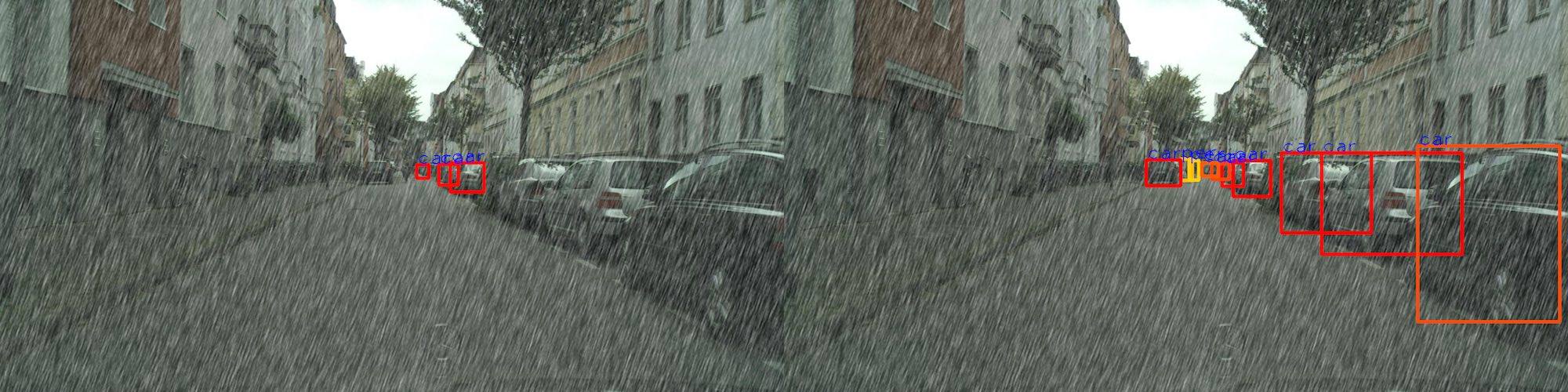}
		\includegraphics[width=0.82\linewidth]{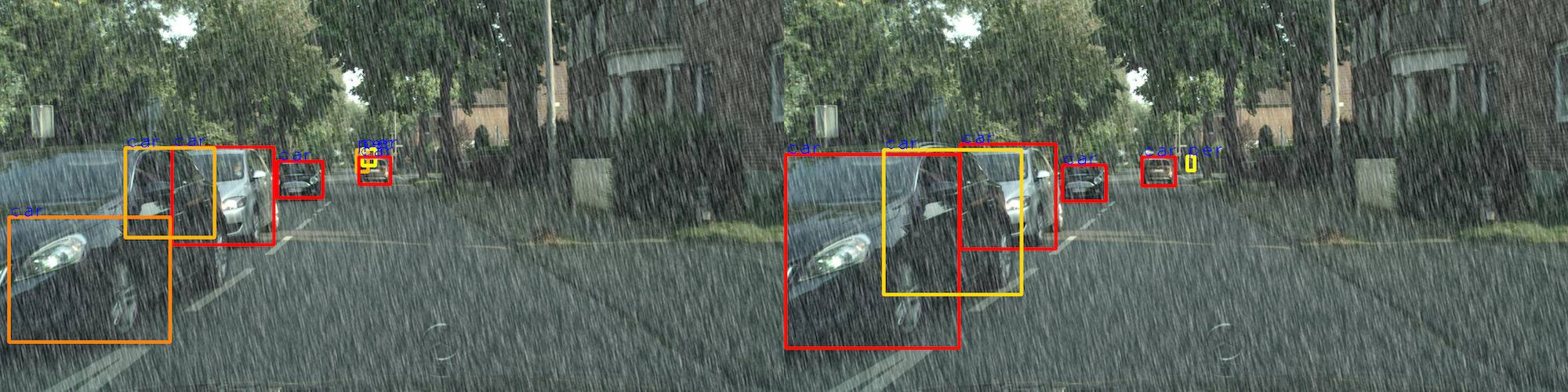}
		\includegraphics[width=0.82\linewidth]{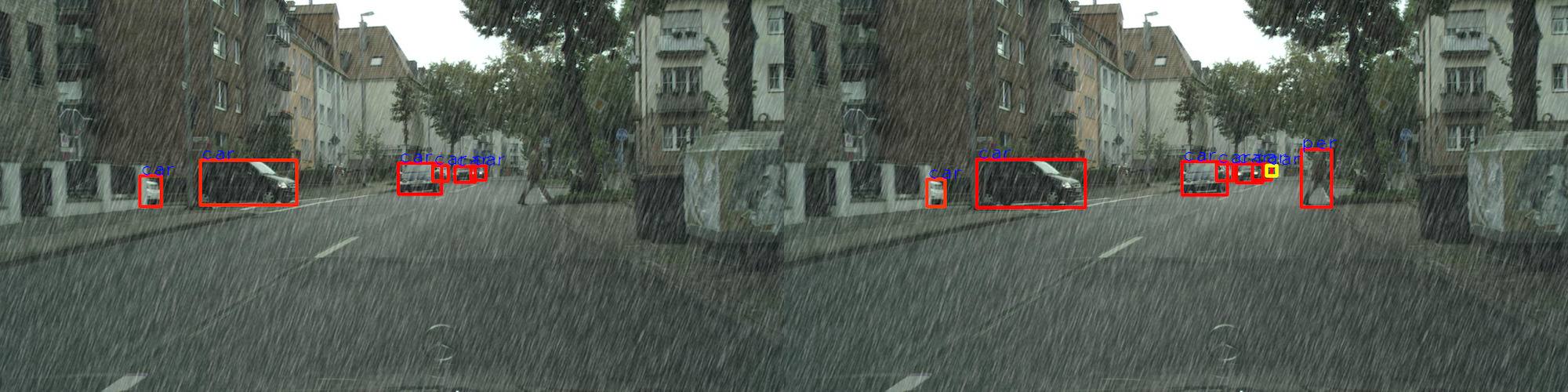}
		\includegraphics[width=0.82\linewidth]{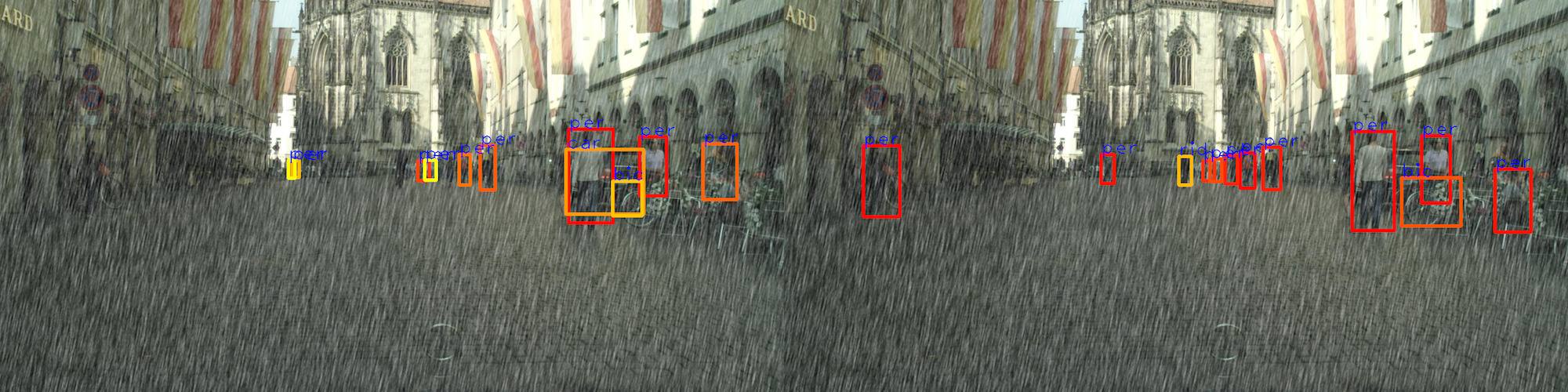}
		\includegraphics[width=0.82\linewidth]{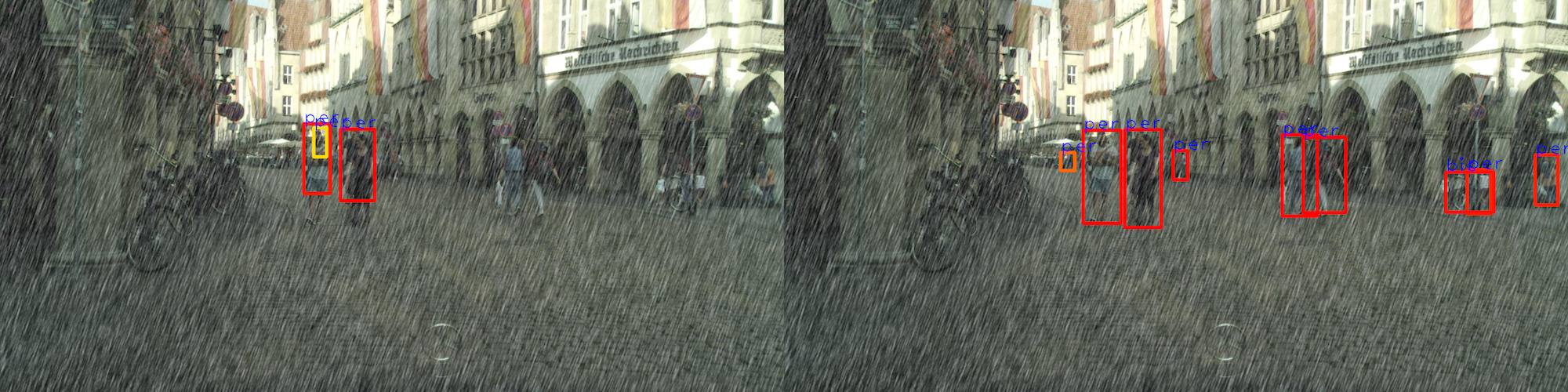}
		\includegraphics[width=0.82\linewidth]{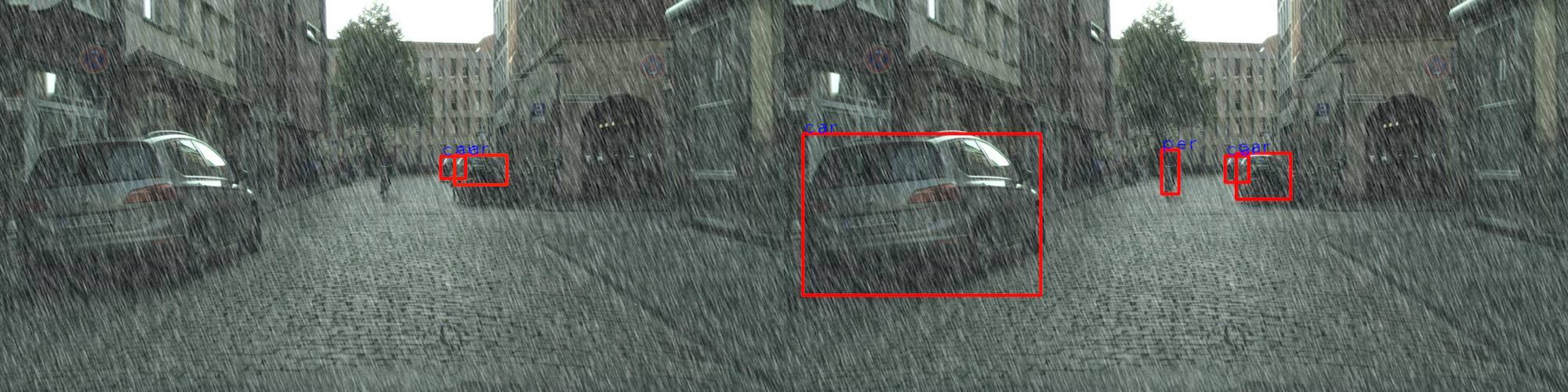}\\
		\includegraphics[width=.25\linewidth]{figures/map}\\
		\vskip -10pt
		(a) \hskip 120pt (b)
	\end{center}
	\vskip -15pt \caption{Detection results on Rainy-Cityscapes. (a) DA-Faster RCNN \cite{Chen2018DomainAF} (b) Proposed method. The bounding boxes are colored based on the detector confidence using the color map as shown. As we can see from the figures, the proposed method is able to produce high confidence predictions and is able to detect more objects in the image.}
	\label{fig:rn}
\end{figure*}

\pagebreak
\noindent\textbf{Cityscapes $\rightarrow$ RTTS}\\

\begin{figure*}[h!]
	\begin{center}
		\includegraphics[width=0.60\linewidth]{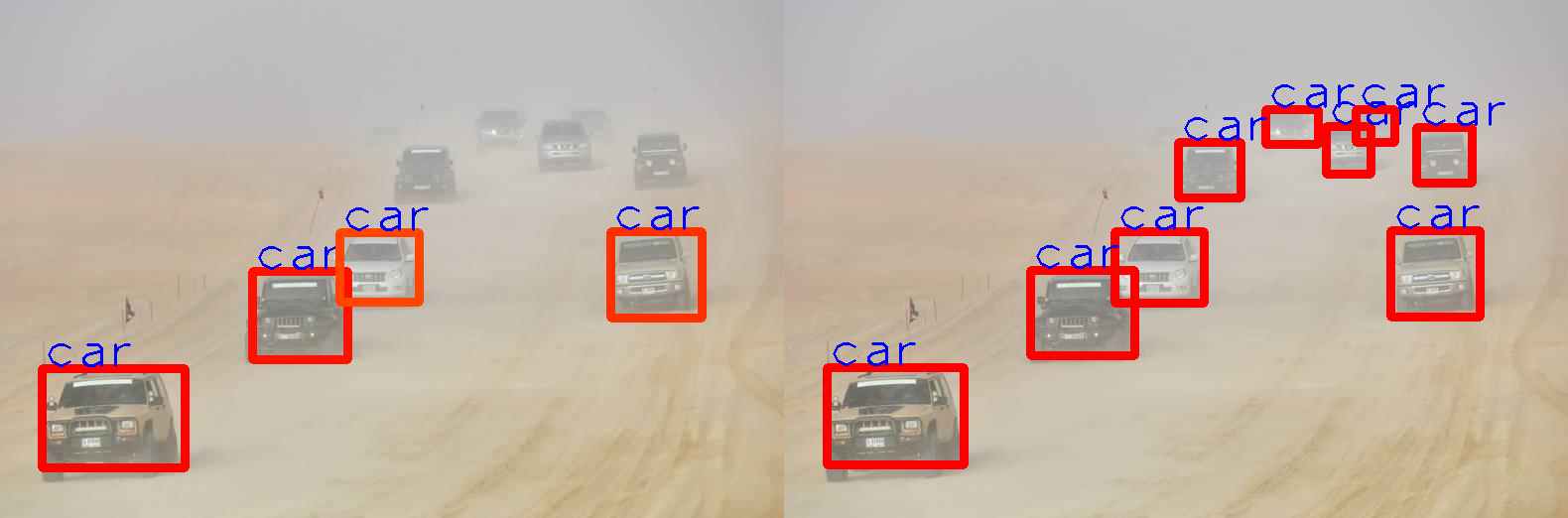}
		\includegraphics[width=0.60\linewidth]{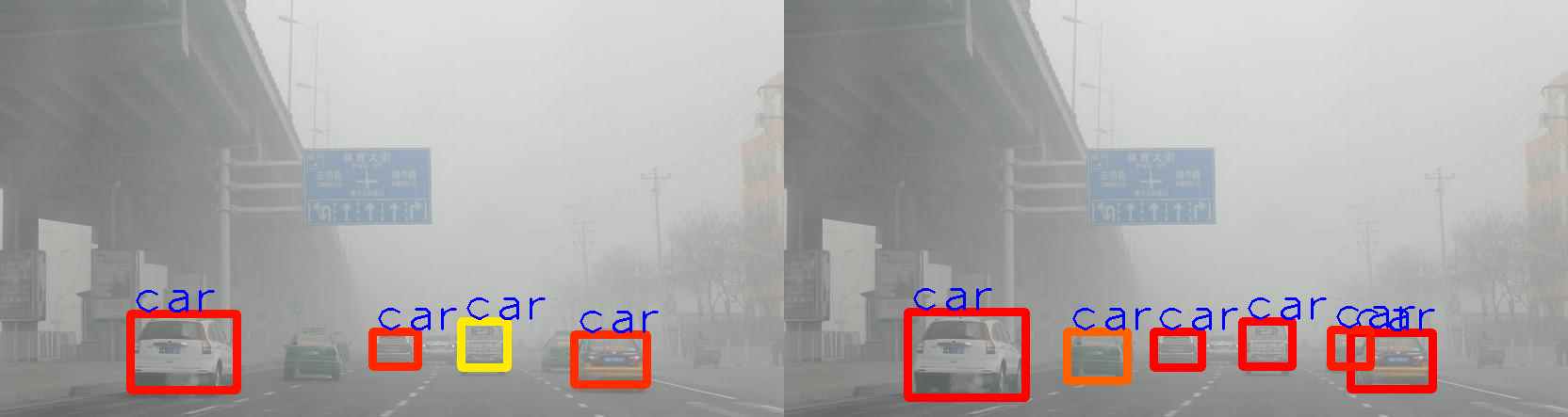}
		\includegraphics[width=0.60\linewidth]{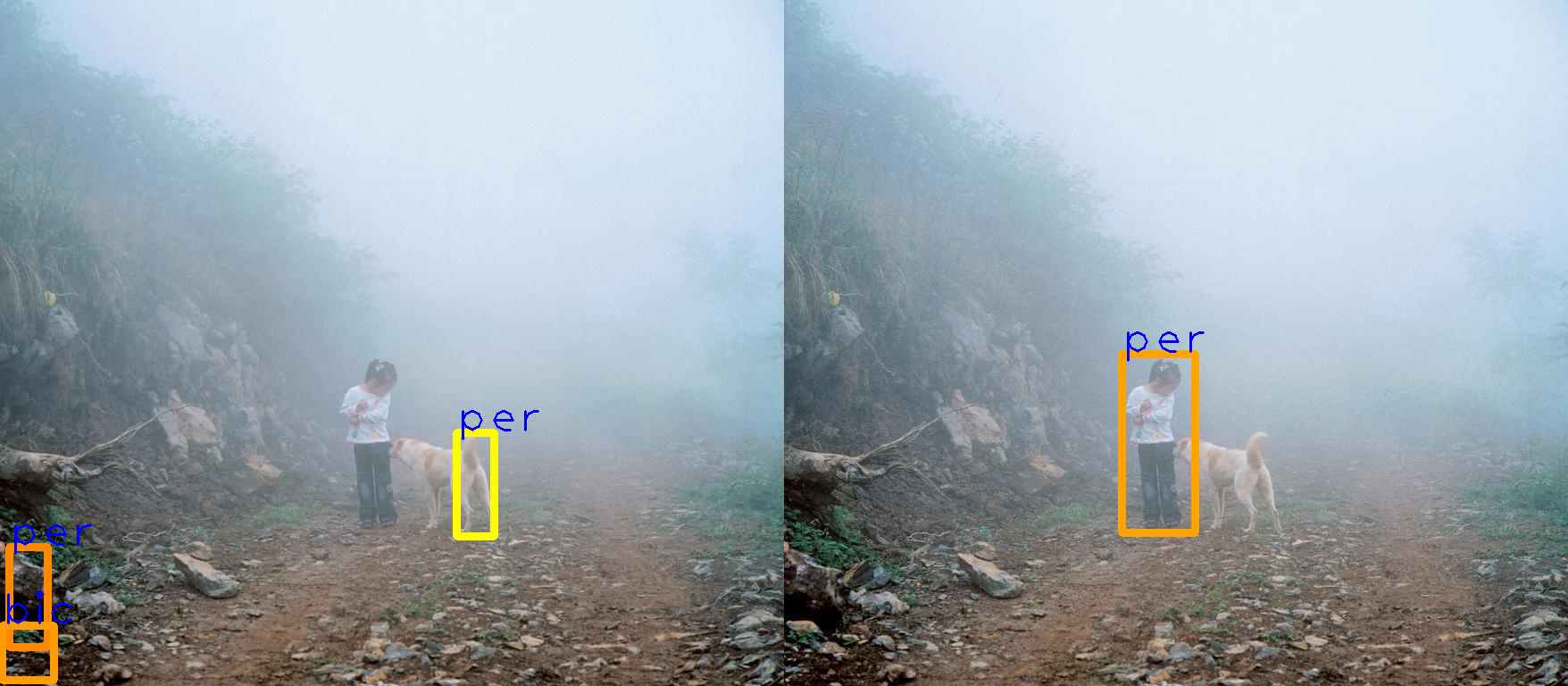}
		\includegraphics[width=0.60\linewidth]{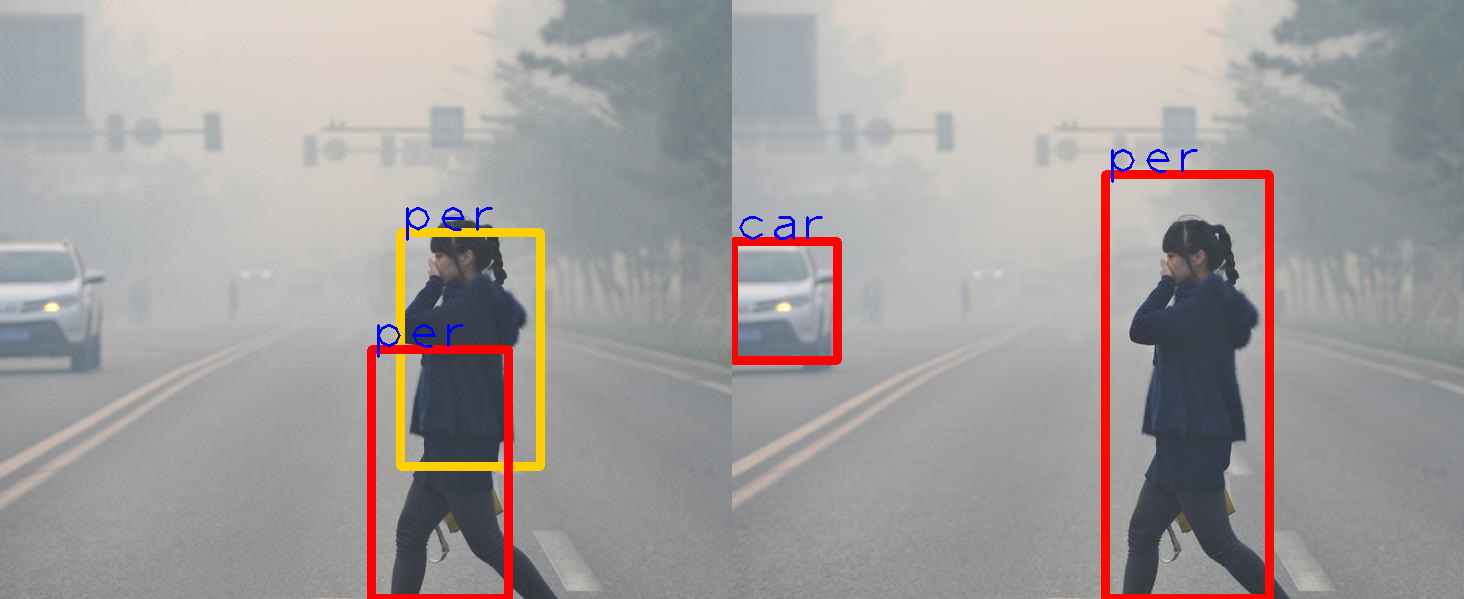}
		\includegraphics[width=0.60\linewidth]{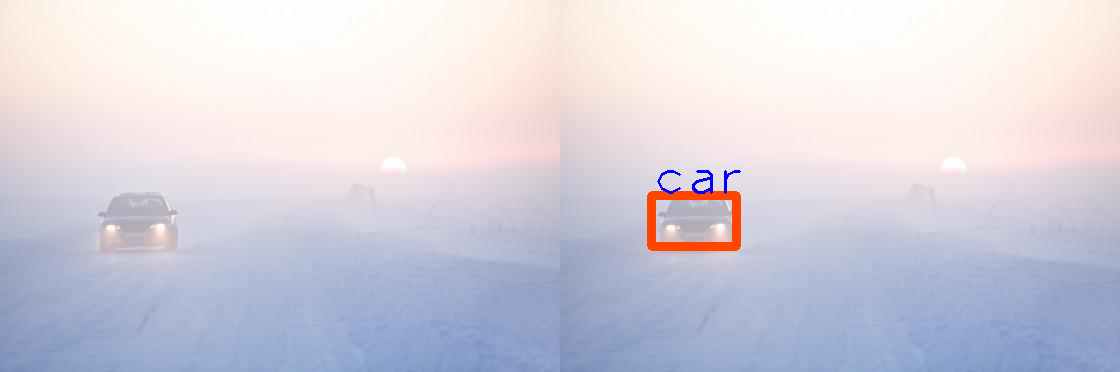}
		\includegraphics[width=0.60\linewidth]{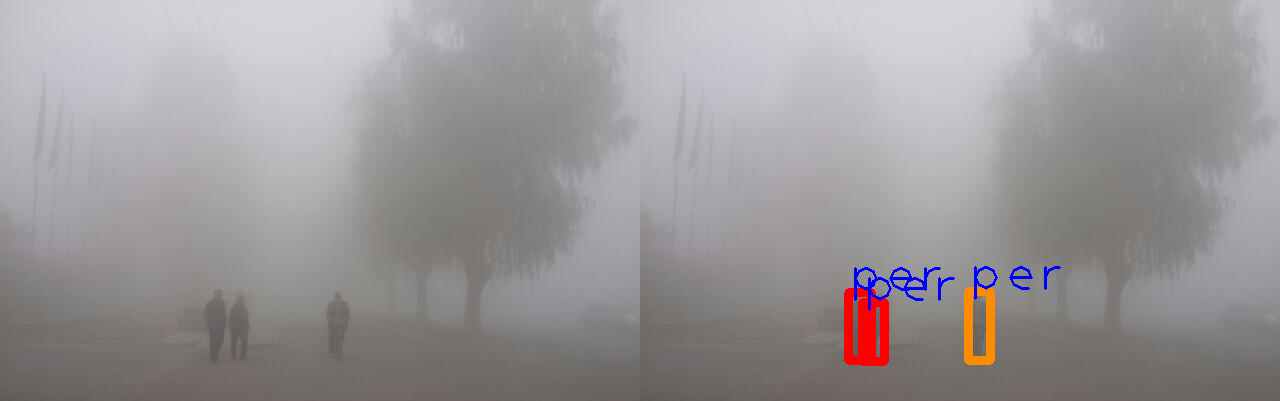}\\
		\includegraphics[width=.25\linewidth]{figures/map}\\
		\vskip -10pt
		(a) \hskip 100pt (b)
	\end{center}
	\vskip -15pt \caption{Detection results on RTTS Dataset. (a) DA-Faster RCNN \cite{Chen2018DomainAF} (b) Proposed method. The bounding boxes are colored based on the detector confidence using the color map as shown. As we can see from the figures, the proposed method is able to produce high confidence predictions and is able to detect more objects in the image.}
	\label{fig:rtts}
\end{figure*}

\pagebreak

\subsection{Rainy-Cityscapes}

We use a subset of 3475 images from Cityscapes to create synthetic rain dataset. Using \cite{synth-rain},  several  masks containing artificial rain streaks are synthesized. The rain streaks are created using different Gaussian noise levels and multiple rotation angles between $70^\circ$ and  $110^\circ$. Next, for every image in the subset of the Cityscapes dataset, we pick a random rain mask and blend it onto the image to generate the synthetic rainy image. More details and example images are provided in supplementary material.

Fig. \ref{fig:rs_examples} shows few sample image examples from Rainy-Cityscapes dataset introduced in the paper.
\begin{figure*}[h!]
	\begin{center}
		\includegraphics[width=0.32\linewidth]{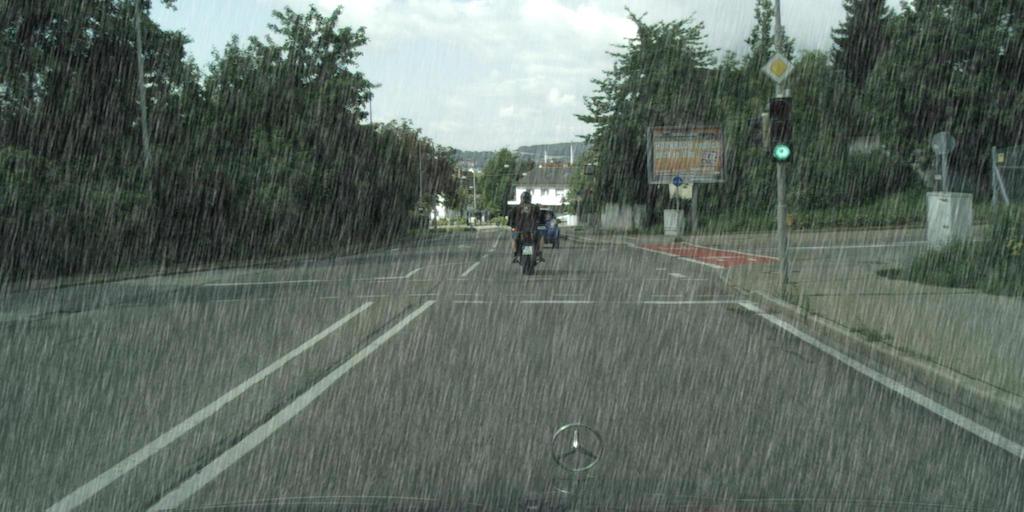}
		\includegraphics[width=0.32\linewidth]{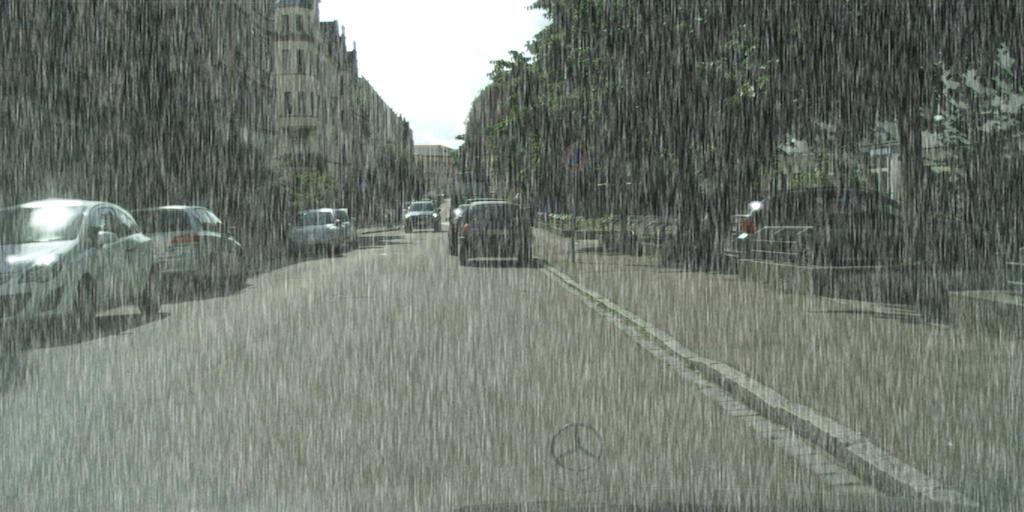}
		\includegraphics[width=0.32\linewidth]{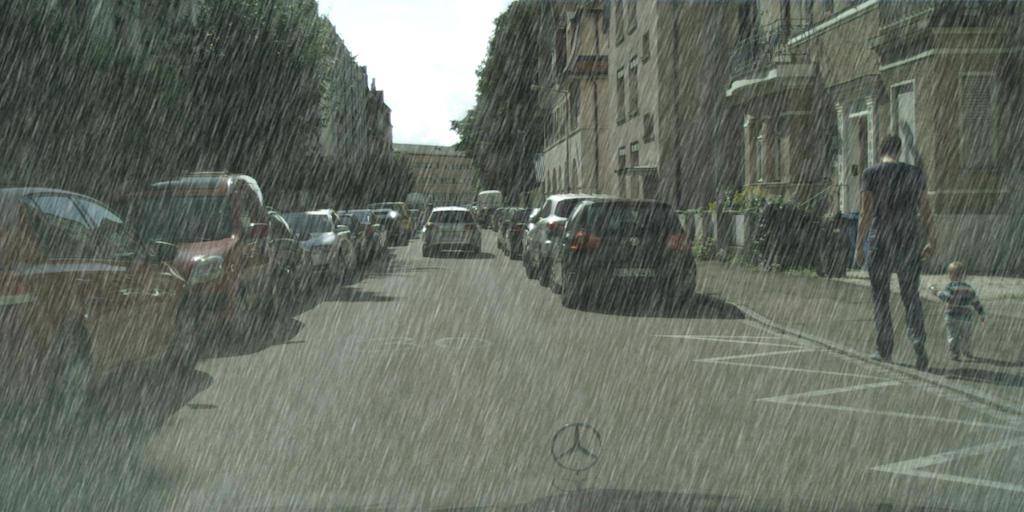}
		\includegraphics[width=0.32\linewidth]{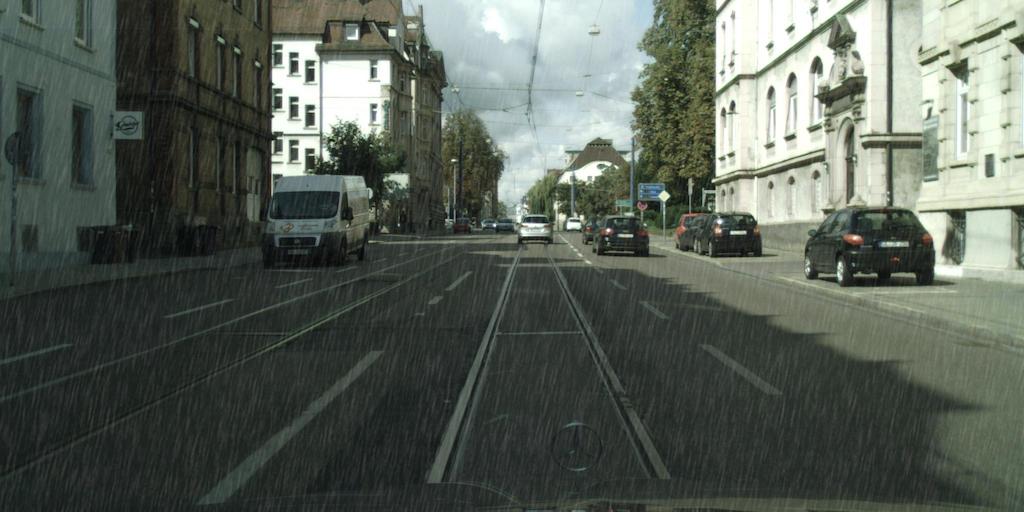}
		\includegraphics[width=0.32\linewidth]{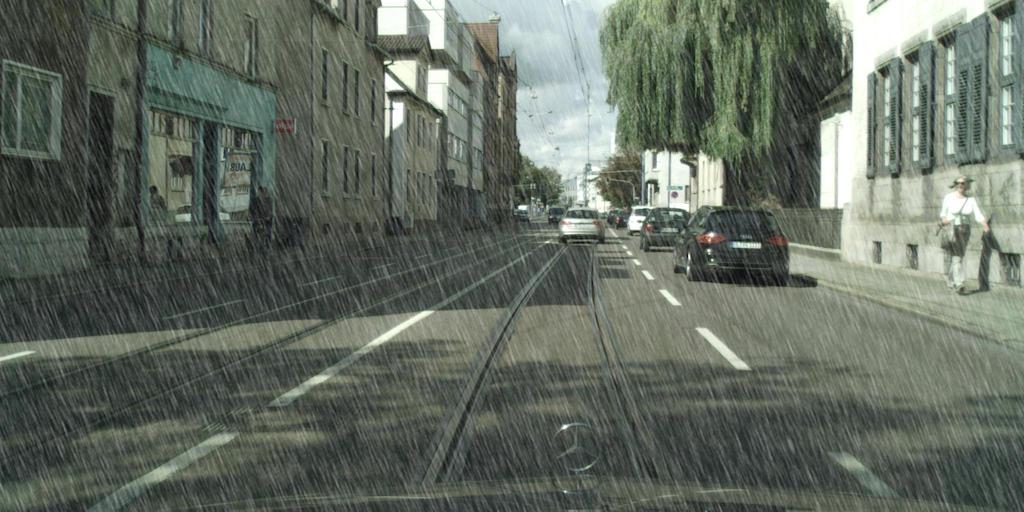}
		\includegraphics[width=0.32\linewidth]{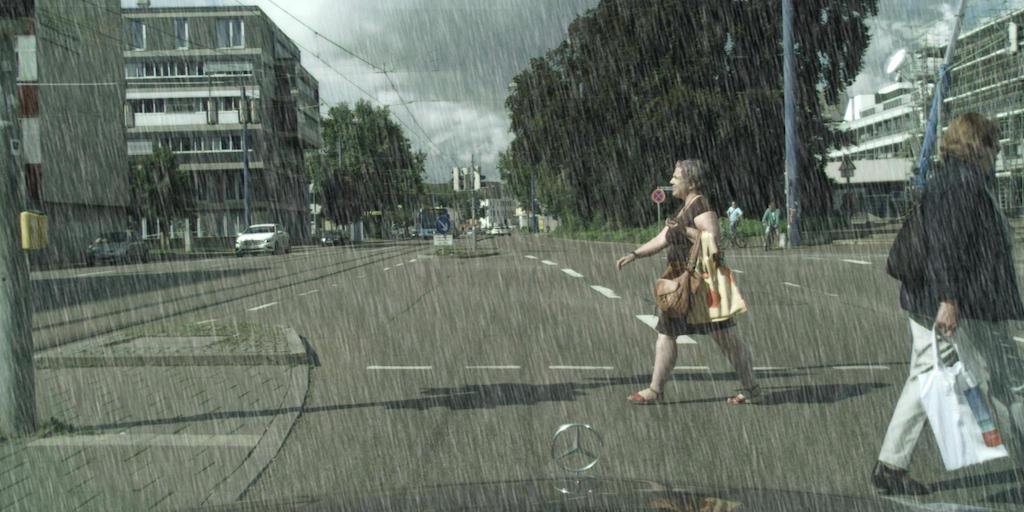}
	\end{center}
	\vskip -15.0pt \caption{Sample images from the Rainy-Cityscapes dataset. }
	\label{fig:rs_examples}
\end{figure*}

\subsection{Preliminary Study : \normalsize{Domain Adaptive Detection in Snow}}\label{sec:snow}

Similar to rain, a snow image can be considered as superposition of clean image and snow-residues. If we denote snowy image as $I_{snow}$, clean image as $I_{clean}$ and snow-residues as $I_{snow-res}$, then snow image can be mathematically written as,
$$I_{snow}=I_{clean}+I_{snow-res},$$
Here, $I_{snow-res}$ can be used as a prior and can be extracted from the snowy image with the help of GMM, similar to the case of rain. We use this model to perform a preliminary experiment on snowy conditions in Sec.~\ref{sec:snow}.

Following the above mention model, we present a preliminary study of domain adaptive detection in snowy conditions. For the experiment, we consider adaptation scenario WIDER-Face $\rightarrow$ UFDD-Snow, which adapts the detection network from labeled clean image dataset, WIDER-Face \cite{yang2016wider}, to unlabeled snow affected image dataset, UFDD-Snow \cite{nada2018pushing}. We compare with the state of art method SWDA \cite{Saito2018StrongWeakDA}. For proposed approach we use the GMM prior as snow weather prior to extract snow residues as explained in the Sec.~\ref{sec:other_weather}. For both methods we use VGG16 \cite{simonyan2014very} as the backbone of the detection network. As we can see from the Table~\ref{tab:snow} that proposed method is able to improve $\sim$10\% over the Faster-RCNN baseline and $\sim$3\% the method SWDA.

\begin{table}[h!]
	\centering
	\caption{Results of the adaptation experiments from WIDER-Face $\rightarrow$ UFDD-Snow.}
	\label{tab:snow}
	\vskip -8pt 
	\resizebox{0.25 \linewidth}{!}{
		\begin{tabular}{|l c|}
			\hline
			Method          & mAP \\ \hline\hline
			FRCNN \cite{ren2015faster}     & 52.1\\
			SWDA \cite{Saito2018StrongWeakDA}  & 58.7\\ \hline \hline
			Proposed  & \textbf{61.9} \\ \hline
		\end{tabular}
	}
\end{table}

\subsection{Extending to Other Weather Conditions}\label{sec:other_weather}

Many other weather conditions have been researched in the literature and have a mathematical model based on the physics of image formation \cite{han2018single}, \cite{you2015adherent}, \cite{you2015adherent}. Our method can be easily extensible to other conditions by utilizing these mathematical models. Here we provide some examples of the prior that can be used based on their corresponding mathematical models:\\

\noindent\textbf{1. Low-light/Sunshine:} Any image can be modeled with the help of Luminance (L) - Reflectance (R) model \cite{mathworks}. This model follows an additive formula in the logarithm and can be written as,
$$log(I) = log(L) + log(R),$$
Here, $I$ is the Image, $L$ is luminance map and $R$ is the reflectance map. Considering this model, for low-light/sunshine conditions, luminance map can be used as a prior information and homoporphic filtering \cite{mathworks} can be used to extract luminance map from the image.\\ \\
\noindent\textbf{2. Water puddle:} The water puddle model has been extensively discussed in \cite{han2018single}. In the mathematical formulation provided by \cite{han2018single}, we can use reflection attention as a prior. The details regarding how to extract the reflectance attention from the image is provided in \cite{han2018single}. \\ \\
\noindent\textbf{3. Adherent water drops:} A detailed discussion adherent water drops mathematical model based on physics of water-droplets is provided in \cite{you2015adherent}. In the model formulation, the term  $I_r$ (Eq. 7  \cite{you2015adherent}) can be used as prior.\\

For the paper we focused mainly on rainy and hazy conditions. However, in future we plan to study the above mentioned weather conditions with the help of priors extracted from the corresponding mathematical model as discussed above.

\end{document}